\pdfoutput=1

\documentclass[11pt]{article}

\usepackage[]{acl}

\usepackage{times}
\usepackage{latexsym}
\usepackage{graphicx}

\usepackage[T1]{fontenc}

\usepackage[utf8]{inputenc}

\usepackage{microtype}

\usepackage{inconsolata}

\usepackage{hyperref}
\usepackage{refstyle}
\usepackage{amsmath}
\usepackage{cleveref}

\usepackage{booktabs}
\usepackage{multirow} %
\usepackage{soul}%
\usepackage{tabularx}
\usepackage{enumitem}

\usepackage{amssymb}

\usepackage{pifont}

\usepackage{graphicx}

\crefformat{section}{\S#2#1#3}
\crefformat{subsection}{\S#2#1#3}
\crefformat{subsubsection}{\S#2#1#3}
\crefrangeformat{section}{\S#3#1#4 to~\S#5#2#6}
\crefmultiformat{section}{\S#2#1#3}{ and~\S#2#1#3}{, #2#1#3}{ and~#2#1#3}
\Crefformat{figure}{#2Fig.~#1#3}
\Crefmultiformat{figure}{Figs.~#2#1#3}{ and~#2#1#3}{, #2#1#3}{ and~#2#1#3}
\Crefformat{table}{#2Tab.~#1#3}
\Crefmultiformat{table}{Tabs.~#2#1#3}{ and~#2#1#3}{, #2#1#3}{ and~#2#1#3}
\Crefformat{appendix}{#2Appx.~\S#1#3}
\crefformat{algorithm}{Alg.~#2#1#3}
\Crefformat{equation}{#2Eq.~#1#3}

\newcommand{\task}{\textsc{1MKR}}

\title{Promote, Suppress, Iterate: How Language Models Answer \\One-to-Many Factual Queries}

\author{
Tianyi Lorena Yan \and Robin Jia \\
University of Southern California \\
\texttt{\{tianyi.yan, robinjia\}@usc.edu}
}

\begin{document}
\maketitle

\begin{abstract} 

To answer one-to-many factual queries (e.g., listing cities of a country), a language model (LM) must simultaneously recall knowledge and avoid repeating previous answers. How are these two subtasks implemented and integrated internally? Across multiple datasets, models, and prompt templates, we identify a promote-then-suppress mechanism: the model first recalls all answers, and then suppresses previously generated ones. Specifically, LMs use both the subject and previous answer tokens to perform knowledge recall, with attention propagating subject information and MLPs promoting the answers. Then, attention attends to and suppresses previous answer tokens, while MLPs amplify the suppression signal. Our mechanism is corroborated by extensive experimental evidence: in addition to using early decoding and causal tracing, we analyze how components use different tokens by introducing both \emph{Token Lens}, which decodes aggregated attention updates from specified tokens, and a knockout method that analyzes changes in MLP outputs after removing attention to specified tokens. Overall, we provide new insights into how LMs' internal components interact with different input tokens to support complex factual recall. \footnote{Code is available at \url{https://github.com/Lorenayannnnn/how-lms-answer-one-to-many-factual-queries}.}

\end{abstract}
\section{Introduction}
\label{1_introduction}

\begin{figure}
    \centering
    \includegraphics[width=1\linewidth]{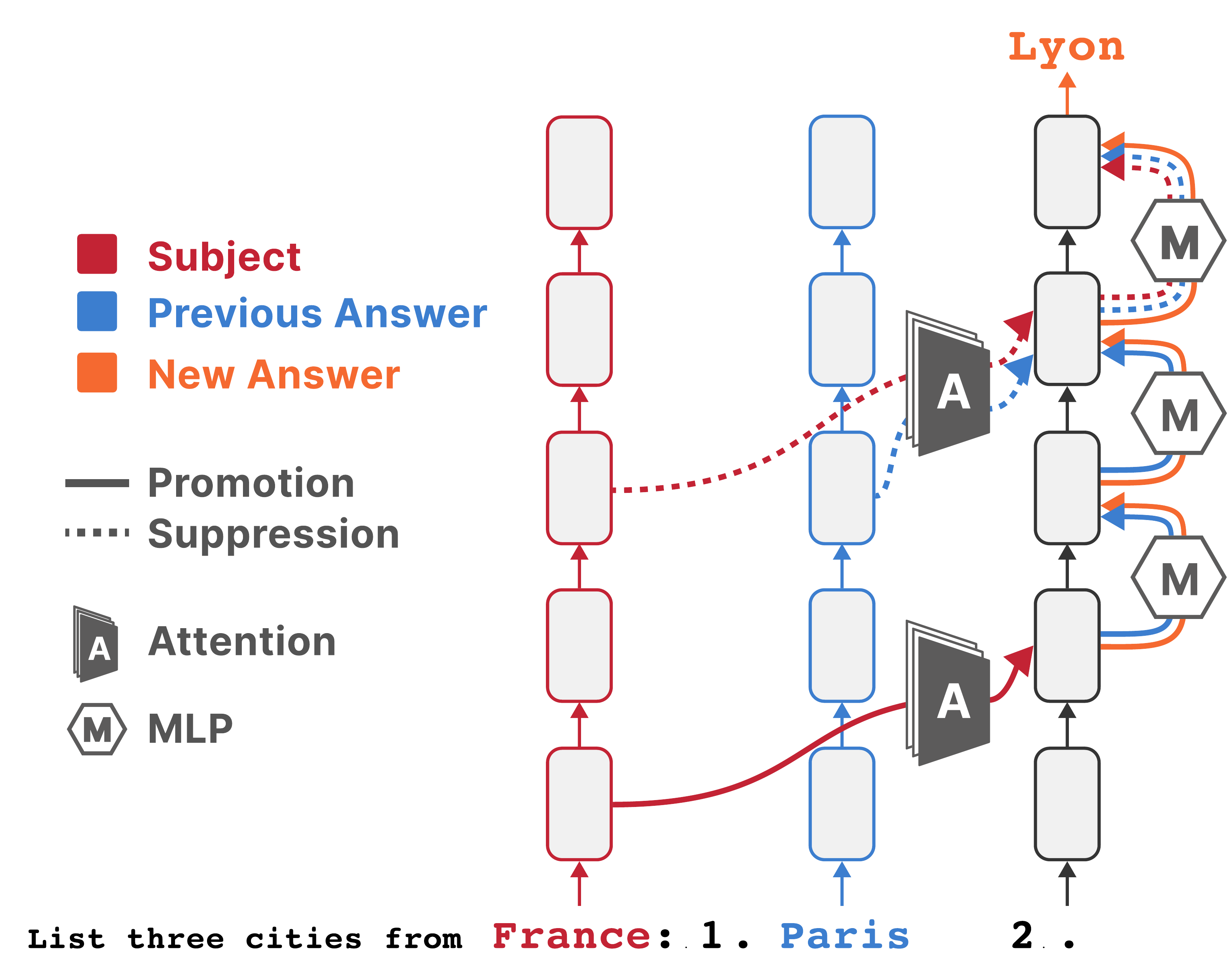}
    \caption{To answer one-to-many factual queries, we found that LMs first use attention to propagate subject information to the last token, which is used by MLPs to promote all possible answers. Attention then attends to and suppresses the subject and previous answer tokens, while MLPs amplify the suppression and further promote new answers.}
    \label{fig:figure_1}
    \vspace{-5mm}
\end{figure}

Transformer-based language models (LMs) store a vast amount of factual knowledge in their parameters \citep{petroni2019language, dai2021knowledge, geva2022transformer}. Many recent works have studied where and how LMs recall this knowledge for one-to-one factual queries, which ask the model to recall a single fact (e.g., the capital of a country) given a subject-relation pair \citep{meng2022locating, geva2023dissecting, merullo2023language}.

In this work, we study the comparatively unexplored task of one-to-many knowledge recall (\task), in which the model must generate a list of answers without repetition. Many real-world relations, such as a country's cities or an artist's songs, are one-to-many. This more complex task requires LMs to integrate multiple pieces of contextual information, including the subject and previously generated answers, to simultaneously perform two subtasks: \textbf{knowledge recall} and \textbf{repetition avoidance}. We uncover LMs' mechanism for \task\ by understanding (1) the overall process by which they generate distinct answers at different steps, and (2) how they perform both answer promotion and repetition avoidance.

To understand the overall process, we early decode \citep{Nostalgebraist_2020} the output of attention and MLPs to examine how the logits of the subject and answer tokens change across layers. We find that LMs first promote all answers and then suppress the ones that have been previously generated. Specifically, attention copies the subject information at the middle layers and MLPs promote all possible answers. Then, both components suppress previous answer tokens at late layers. These observations hold for both Llama-3-8B-Instruct and Mistral-7B-Instruct-v0.2 across three datasets.
 
To examine how LMs implement knowledge recall and repetition avoidance, we first run causal tracing \cite{meng2022locating} to locate tokens that are critical to LMs' outputs; these important tokens include the subject, previous answers, and the last token. Then, we analyze how both attention and MLP layers use these tokens. For attention, we propose Token Lens, a new technique that aggregates and then unembeds the results of attending to a given token or span; in this way, we can observe how attention to each token promotes or suppresses different output tokens. For MLPs, we design an attention knockout method inspired by \citet{geva2023dissecting}: we knock out the attention from the last token to the target tokens and examine the resulting change in MLP output logits to determine how MLPs use target token information.
We find that LMs use both the subject and previous answer tokens for knowledge recall: attention propagates the subject information from the subject to the last token, and MLPs leverage the information and previous answer tokens to promote answers. In addition, previous answer tokens trigger suppression of themselves: attention attends to and suppresses previous answer tokens, while MLPs amplify the suppression signal. LMs aggregate this information at the last token to generate distinct answers across steps. 

Overall, our study elucidates how LMs use attention and MLPs to interact with different tokens and perform knowledge recall and repetition avoidance for \task. We hope this work opens pathways for analyzing more complex tasks requiring dynamic integration of contextual information.

\section{Related Work}
\label{2_related_work}

\paragraph{Interpretability of Language Models.} 
Works on mechanistic interpretability aim to reveal the function of different components in LMs \citep{elhage2021mathematical, bansal2022rethinking}, such as neurons \citep{dai2021knowledge, gurnee2023finding}, attention heads \citep{michel2019sixteen, olsson2022context}, and MLPs \citep{geva2020transformer, geva2022transformer}. 
In particular, how LMs store and use knowledge has been widely studied by many prior works \citep{petroni2019language, bouraoui2020inducing, cao2021knowledgeable, dalvi2022discovering, da2021analyzing}. 
However, most prior studies have mainly focused on one-to-one knowledge recall, where LMs retrieve a single fact given a subject-relation pair. In this work, we study how LMs' components contribute to one-to-many knowledge recall, which is a more complex setting that requires LMs to integrate multiple types of contextual information: subject, relation, and previously generated answers.

\paragraph{Attribution Methods.} 
Prior works have introduced various methods for analyzing the function of different components, including probing \citep{burns2022discovering, li2024inference}, patching \citep{goldowsky2023localizing, ghandeharioun2024patchscope}, early decoding \citep{Nostalgebraist_2020, merullo2023language}, and knocking out component outputs to assess their impact on models' outputs \citep{chang2023localization, li2023circuit, geva2023dissecting}. Our method, \hyperref[6_1_1_attn_token_lens]{Token Lens} and \hyperref[6_1_2_mlps_attn_knockout]{attention knockout} (inspired by \citet{geva2023dissecting}) examines the importance of attention and MLPs by early decoding their token-level outputs, revealing how LMs use the two components to integrate information from various parts of the input.

\paragraph{Dissecting Component Functions.} 
Recent works have studied the functions of MLPs and attention in knowledge recall given subject-relation pairs. \citet{meng2022locating} and \citet{geva2023dissecting} demonstrate that MLPs enrich subject representations at early layers, while \citet{geva2022transformer} and \citet{merullo2023language} highlight how MLPs promote correct answer tokens by writing updates to the residual stream and adjusting the vocabulary probabilities. This mechanism is still essential for the model to generate multiple answers tied to the given subject. Prior works have also shown that attention and MLPs play a key role in extracting important tokens and suppressing repeated ones \citep{wang2022interpretability, mcdougall2023copy, voita2023neurons, merullo2023circuit, tigges2024llm}, which is essential for preventing the model from generating duplicate answers. \citet{merullo2024talking} further decomposes attention heads and identifies low-rank subspaces in which components communicate to selectively inhibit repetitive items from a list given in the context, which also involves list processing and repetition avoidance but not recalling factual knowledge from model parameters.

\section{Problem Settings}
We first introduce the task of one-to-many knowledge recall and describe our experiment settings. 

\subsection{Task: One-to-Many Knowledge Recall}
\label{3_1_task_one_to_many_knowledge_recall}

In \task, a language model is given a subject entity $s$ and a relation $r$, and must generate a set of corresponding object entities $O = \{o^{(1)}, o^{(2)}, \dotsc,o^{(n)}\}$ that are related to $s$ through $r$. All generated object entities must be distinct, that is, $o^{(i)} \neq o^{(j)}$ for $i \neq j$. For example, given $s=\text{"U.S.A."}$ and $r=\text{"cities of"}$, one possible valid set of object entities is $O= \{\text{Los Angeles}, \text{San Francisco}, \text{Seattle}\}$. To perform this task, the model must perform two key subtasks:
\vspace{-3mm}
\begin{enumerate}[wide, labelwidth=!, labelindent=0pt]
    \item \textbf{Knowledge recall}: The model must identify and extract the subject $s$ from the input and retrieve entities that are connected to $s$ through the relation $r$ from its internal knowledge.
    \vspace{-3mm}
    \item \textbf{Repetition avoidance}: The model must not generate duplicate entities.
    \vspace{-1mm}
\end{enumerate}

\paragraph{Possible mechanisms.}
Multiple different mechanisms could be used by the model to perform \task.
On one hand, the model could use different attention heads to promote a different answer at each timestep. 
It could first use suppression heads \citep{wang2022interpretability} to identify previously generated answers, then change the attention patterns of subsequent heads to avoid promoting those answers.
Such a mechanism would mirror the use of suppression heads to avoid generating incorrect, repetitive tokens in the IOI task \citep{wang2022interpretability}.
To promote answers, the model could attend to the subject token position, which could encode different answers in different attention value vectors due to subject enrichment \citep{geva2023dissecting}.

On the other hand, the model could first promote all relevant answers and then suppress previously generated ones. It could extract all possible answers from the subject representation \citep{geva2023dissecting, meng2022locating}, regardless of which object entities have been generated. Then, copy suppression heads could identify previous answer tokens and prevent the model from generating them, similar to \citet{mcdougall2023copy}.
The results of knowledge recall and repetition avoidance could be additively combined in the residual stream to yield a correct and non-duplicate output, similar to \citet{chughtai2024summing}.
In this paper, we uncover the true mechanism that the model uses for one-to-many knowledge recall.

\subsection{Datasets and Models}

\begin{table*}
    {\fontsize{8.5pt}{10pt}\selectfont
    \setlength{\tabcolsep}{2.8pt}
    \centering
    \begin{tabular}{c|cccc|cc}
    \toprule
    \textbf{Dataset} & \textbf{Subject ($s$)} & \textbf{Relation ($r$)} & \textbf{Object ($o$)} & \textbf{\# Entries} & \textbf{Llama-3-8B-Instruct (Acc)} & \textbf{Mistral-7B-Instruct-v0.2 (Acc)} \\
    \midrule
    Country-Cities       & Country              & contains              & Cities    & 168         & 122/168 (72.8\%)            & 118/168 (70.2\%)            \\
    Artist-Songs       & Artist               & performer of          & Songs      & 2077         & 276/2077 (13.3\%)           & 221/2077 (10.6\%)           \\ 
    Actor-Movies       & Actor                & acted in              & Movies      & 8790        & 1235/8790 (14.1\%)          & 799/8790 (9.1\%)          \\
    \bottomrule
    \end{tabular}
    \caption{Models' performance on all three datasets averaged across three prompt templates. Lower accuracy may be attributed to long-tail entities, outdated data, and overly strict exact-match evaluations. Our analysis focuses on the correct cases. See \Cref{appendix_prompts} for the prompt templates and per-template performance.}
    \label{tab:datasets_and_model_performance}}
\end{table*}

We curate three \task\ datasets on different topics: (1) cities of a country,\footnote{\url{https://simplemaps.com/data/world-cities}} (2) songs performed by an artist,\footnote{\url{https://www.kaggle.com/datasets/salvatorerastelli/spotify-and-youtube}}, and (3) movies acted in by an actor or actress.\footnote{\url{https://www.kaggle.com/datasets/darinhawley/imdb-films-by-actor-for-10k-actors}} A summary of the datasets is provided in \Cref{tab:datasets_and_model_performance}. For each dataset, the number of object entities $n=3$.\footnote{We also tested larger $n$. The models fail to generate at least $100$ correct cases except for the Actor-Movies dataset with $n=5$, where all major results align with those discussed in the main section. See \Cref{appendix_more_answer_steps} for details.} We filter out subjects that are associated with fewer than three object entities for the specified relation. 

We study two LMs: Llama-3-8B-Instruct \cite{llama3modelcard} and Mistral-7B-Instruct-v0.2 \cite{mistral7Bmodelcard}. We have three prompt templates for each model and dataset, which are shown in \Cref{appendix_prompts}. To create the data for analyzing LMs' behaviors, we first generate three answers using greedy decoding, ensuring consistent outputs for examining component behaviors across different answer steps. We then retain the entries where all three predicted answers are correct to focus on cases where the models' knowledge is accurate. 

\Cref{tab:datasets_and_model_performance} shows the number of correct predictions made by the models across the datasets. The low accuracy may be explained by (1) long-tail entities (e.g., less popular actors or songs), (2) outdated datasets compared to the model's knowledge, and (3) the strict use of exact match evaluation (e.g., ``Mission: Impossible'' is considered incorrect even if given ``Mission: Impossible - Fallout'' is in the label list). 
For all (dataset, model) pairs, we have at least 100 correct instances, providing a sufficient sample size for the analysis. For the rest of the paper, we focus only on the correct cases. When analyzing models' behaviors at step $i$ ($i = 1, 2, 3$), we keep all tokens before the first token of the $i$th answer as input. Refer to \Cref{appendix_sample_responses_and_analysis_data_example} for examples and details. We report results macro-averaged across all models, datasets, and prompt templates in the main section. Refer to the appendix for full results of all answer steps and specific models and datasets. We run all experiments on a single RTX A6000 GPU.

\section{Decoding the Overall Mechanism}

To understand how LMs perform \task, we first inspect the outputs of attention and MLP across layers. We aim to understand how knowledge recall and copy suppression coordinate to produce different correct answers across generation steps. 

\subsection{Method: Decoding Component Outputs}
\label{4_1_method_decoding_components_outputs}
Given a transformer LM with $L$ layers, each layer $l$ has a multi-headed attention (MHA) and a MLP layer for $l=1\dotsc L$. Let $a^{(l)} \in \mathbb{R}^d$ and $m^{(l)} \in \mathbb{R}^d$ be the outputs of the MHA and MLP at layer $l$ at the last token position\footnote{We focus on the last token position as the model directly uses it to generate the next answer.} respectively. Similar to \citet{Nostalgebraist_2020} and \citet{geva2022transformer}, we (early) decode $a^{(l)}$ and $m^{(l)}$ by passing them through the final layer layernorm and unembedding matrix $U \in \mathbb{R}^{|\text{Vocab}| \times d}$ and obtain the logits to examine their contributions to knowledge recall and repetition avoidance:
    \begin{equation}
        \text{logits} = U \cdot \text{LayerNorm}(z^{(l)})
    \end{equation} 
    where $z^{(l)}$ is $a^{(l)}$ or $m^{(l)}$, and $\text{LayerNorm}(\cdot)$ denotes the final layernorm. In this paper, $\text{LayerNorm}(\cdot)$ is the RMSNorm \cite{zhang2019root}. Note that the RMSNorm is calculated based on the input's hidden state from the final layer, not directly on $a^{(l)}$, ensuring consistent normalization across layers and components \cite{chang2024parts}.

\subsection{LMs Promote Then Suppress}
\label{4_2_LMs_promote_then_suppress}
We analyze the logit values of the first tokens of object entities predicted across three answer steps and the subject. A positive logit indicates promotion, while a negative logit suggests suppression. Our analysis shows that LMs use both attention and MLPs to promote all possible answers at each step while suppressing repetitions.

\begin{figure*}
    \centering
    \includegraphics[width=1\linewidth]{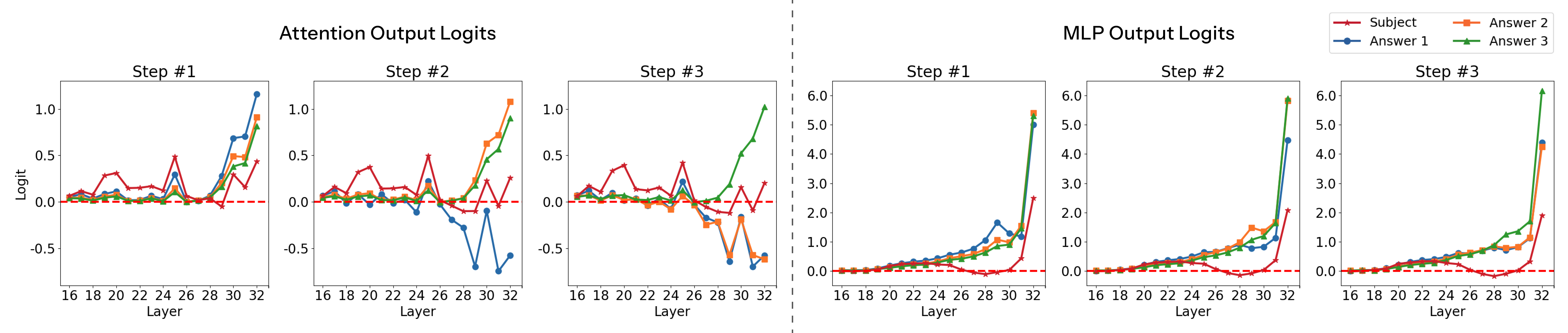}
    \caption{Logit of the subject and answer tokens from unembedding attention and MLP outputs. Attention primarily promotes the subject at the middle layers, then promotes new answers and suppresses previous answers at deeper layers. MLPs consistently promote all answers; at deeper layers, they also decrease the logits of previously generated answers. Early layers are omitted as logits are near zero. See \Cref{appendix_unembed_attn_mlp_outputs_full_figures} for full figures.
    }
    \label{fig:unembed_component_output_logit}
\end{figure*}

\paragraph{Attention primarily copies subject information.} As shown in \Cref{fig:unembed_component_output_logit}, attention outputs positive logits for the subject token in the middle layers across all three answer steps. While the three answers are slightly promoted at layer $25$, their logits are still close to zero and have a smaller magnitude compared to that of the subject at the middle layers. This pattern indicates that attention copies or propagates subject information at the last token position. Interestingly, the answer promotion pattern is more evident in the Country-Cities dataset (\Cref{fig:country_cities-llama-attn_mlp_output_logit}, \Cref{fig:country_cities-mistral-attn_mlp_output_logit}) but not in the Artist-Songs and the Actor-Movies datasets (\Cref{fig:artist_songs-llama-attn_mlp_output_logit}, \Cref{fig:artist_songs-mistral-attn_mlp_output_logit}, \Cref{fig:actor_movies-llama-attn_mlp_output_logit}, \Cref{fig:actor_movies-mistral-attn_mlp_output_logit}). %

\paragraph{MLPs promote all possible answers.} From the middle to later layers, MLPs consistently output positive logits for all three possible answers (\Cref{fig:unembed_component_output_logit}). These logits increase across generation steps, with their magnitude significantly exceeding that of attention logits. These findings suggest that MLPs strongly promote all possible answers regardless of prior predictions, thereby providing a stronger answer promotion signal compared to attention.

\paragraph{Previously generated answers are suppressed at later layers.} Both attention and MLPs suppress answers that have been generated previously. Starting from layer $28$, attention outputs negative logits for $o^{(1)}$ at step $2$ and for both $o^{(1)}$ and $o^{(2)}$ at step $3$ (\Cref{fig:unembed_component_output_logit}). Similarly, MLPs decrease the logit of previous answers at the same layer. Since MLPs themselves cannot attend back to early tokens, this suppression likely results from leveraging suppression signals from attention, a hypothesis further investigated in \Cref{6_3_previous_answer_tokens_for_both_repetition_avoidance_and_knowledge_recall}. \\

\noindent In the final layers, both attention and MLPs increase answers' logits, especially those that have not been generated. This pattern may be explained by how LMs use the final layers to adjust the logits and regulate the confidence or certainty of their predictions \cite{stolfo2024confidence}. Overall, all the observations above demonstrate that LMs promote all three answers and then suppress previously generated ones.

\section{Which Tokens Matter?}
To better understand the promote-then-suppress mechanism, we now investigate how LMs implement knowledge recall and repetition avoidance. In this section, we use causal tracing \cite{meng2022locating} to identify the input tokens that most influence model predictions. In \Cref{6_analyze_critical_tokens}, we analyze how these tokens are used by attention and MLPs to facilitate knowledge recall and repetition avoidance.

\subsection{Which Tokens Should Be Noised?}
\label{5_1_which_tokens_should_be_noised}
Prior work shows that in order to recall knowledge, LMs encode information about relevant object entities in subject tokens and retrieve this information via attention
\citep{geva2023dissecting, meng2022locating}. Other work shows that LMs avoid repetition by using attention heads to attend to previous tokens and suppress them \citep{mcdougall2023copy, wang2022interpretability, merullo2023circuit}. %
Thus, we hypothesize that the subject and previous answer tokens play decisive roles in our two key sub-tasks (\Cref{3_1_task_one_to_many_knowledge_recall}).

To confirm these hypotheses, we use causal tracing \cite{meng2022locating}: we separately add noise to the subject and previous answer tokens, restore selected components' activations to their values without noise, and visualize the difference in the probability of $o^{(i)}$ that will be predicted at each answer step $i$ before and after the restoration. This approach allows us to measure the impact of specific token activations on the models' outputs.

\begin{figure}[h!]
    \centering
    \includegraphics[width=1\linewidth]{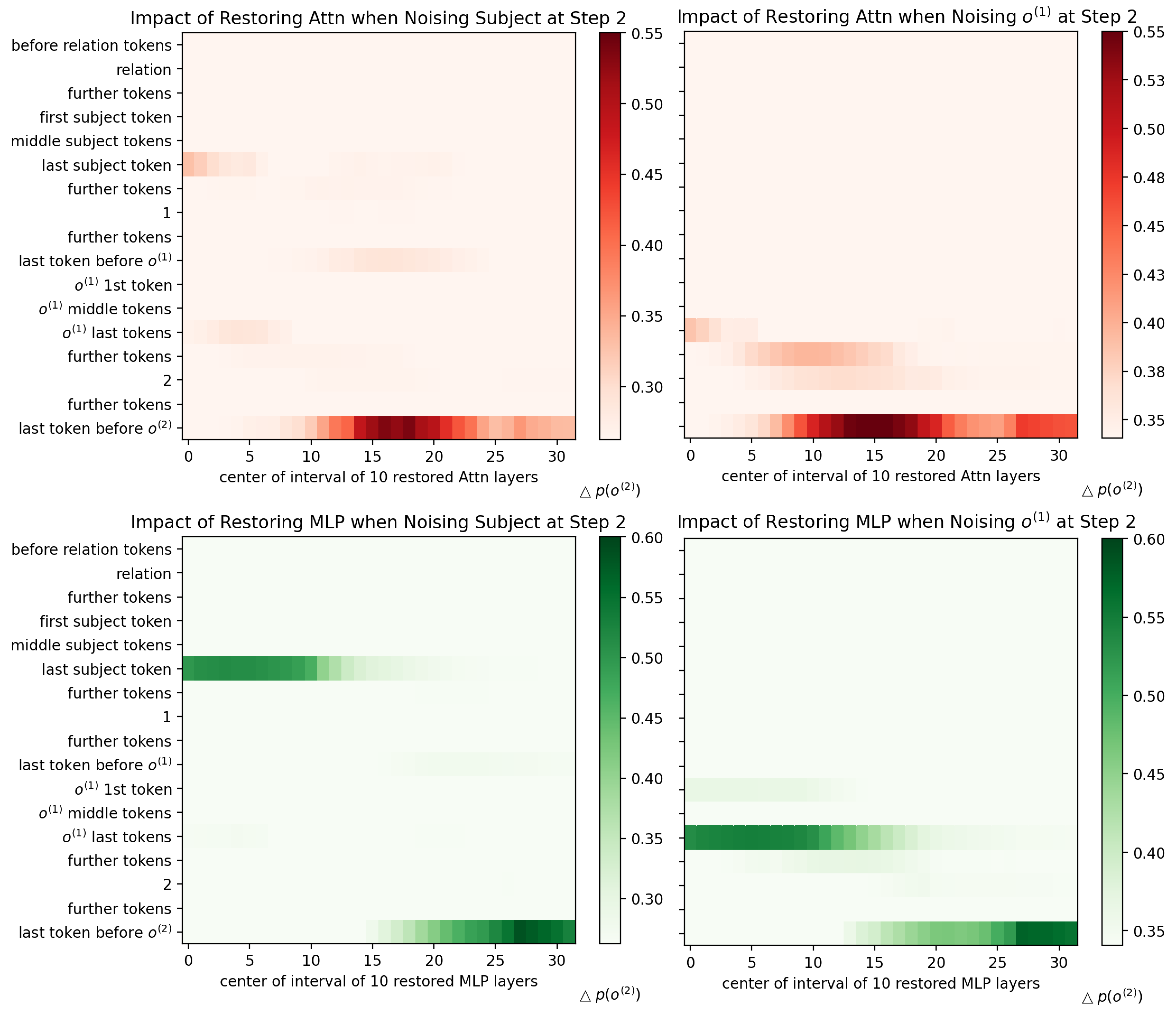}
    \caption{The impact of attention and MLPs' activations on LMs' predictions when intervening on the subject (left) and previous answer tokens (right) at step $2$. %
    The probability differences all peak around or above $0.55$, reflecting the importance of both the subject and previous answer tokens. See \Cref{appendix_causal_tracing_figures} for figures of other answer steps, which have similar patterns.
    }
    \label{fig:causal_tracing_for_main}
    \vspace{-5mm}
\end{figure}

\paragraph{Intervention on Subject.} \Cref{fig:causal_tracing_for_main} visualizes the impact of attention and MLPs on LMs' predictions when intervening on the subject tokens at step $2$ (Refer to \Cref{appendix_causal_tracing_figures} for figures of other answer steps and specific models and datasets, which have similar patterns). The probability difference peaks around or above $0.55$ for both components, confirming our hypothesis that the subject plays a crucial role in knowledge recall. Attention's contributions peak in the middle layers at the last token, while MLPs dominate in early layers at the subject token and in late layers at the last token. These observations suggest that attention propagates subject information from early MLP layers to the last token, where MLPs may leverage it for answer promotion, as discussed in \Cref{6_2_subject_tokens_for_knowledge_recall}.

\paragraph{Intervention on previous answers.} Noising previous answer tokens also leads to high probability changes in LMs' output probabilities, with an average difference of around or above $0.55$ across answer steps (\Cref{fig:causal_tracing_for_main}). This finding supports our hypothesis that previous answer tokens are also critical to LMs' outputs. Similar to the results of noising the subject, attention's contributions peak in both the middle and the last layers at the last token. MLPs dominate in early layers at the previous answer positions and in late layers at the last token, reflecting that the previous answer tokens are used by both components to make nontrivial contributions to models' predictions.

\section{Analyze Critical Tokens}
\label{6_analyze_critical_tokens}

The causal tracing analysis confirms that both the subject and previous answer tokens are important for handling one-to-many factual queries. To determine whether the subject primarily supports knowledge recall and previous answer tokens drive suppression, we next analyze how attention and MLPs utilize these tokens, as well as the last token that the model uses to predict the next answer.

\subsection{Methodology for Analyzing Tokens}
\label{6_1_token_analysis_methodology}
To analyze how attention and MLPs utilize the subject, previous answer, and last tokens, we develop techniques to unembed their token-specific outputs and examine their roles in knowledge recall and suppression.

\paragraph{Attention: Token Lens.}
\label{6_1_1_attn_token_lens}
For attention, we propose Token Lens, a new technique that unembeds the aggregated outputs of attention to specified tokens.
Let $t=\{t_1, ... t_k\}$ denote the target tokens we are examining. $t$ can be the subject $s$, an object entity answer $o^{(i)}$, or the last token of the input. %
Let $a^{(l_i)}$ be the $i$th attention head in layer $l$ of a transformer LM, for $i=1\dotsc n$ and $l=1\dotsc L$. Let $p^{(l_i)}_{t_j} \in \mathbb{R}$ denotes $a^{(l_i)}$'s attention weight between the last input token\footnote{We only need to do the analysis when LLMs start to generate the next answer. Therefore, we are only looking at the last token of the input.} and the $t_j$th token of the input. Similarly, let $v^{(l_i)}_{t_j} \in \mathbb{R}^{d_{\text{head}}}$ denotes the value vector of $a^{(l_i)}$ for the the $t_j$th token.

We first gather the information that each attention head $a^{(l_i)}$ aggregates from all target tokens, which is calculated as the sum of all weighted value vectors of $t$ of $a^{(l_i)}$:
    \begin{equation}
        a^{(l_i)}_e = \sum_{j=1}^{k} p^{(l_i)}_{t_j} \cdot v^{(l_i)}_{t_j}
    \end{equation}
Then, the full attention output of the target tokens from the $l$th layer is:
    \begin{equation}
        a^{(l)}_e = W^{(l)}_o \cdot \text{Concat}(a^{(l_1)}_e, \dotsc, a^{(l_n)}_e)
    \end{equation}
    where $W^{(l)}_o \in \mathbb{R}^{d \times nd_{\text{head}}}$ is the output projection matrix of layer $l$. This vector $a^{(l)}_e \in \mathbb{R}^d$ represents the contribution of MHA at layer $l$ to the output from the target tokens.

Finally, following the same approach of (early) decoding attention and MLP outputs in \Cref{4_1_method_decoding_components_outputs}, we unembed $a^{(l)}_e$ to obtain the logits of the first token of the subject and answers and examine how attention uses the target tokens to perform promotion or suppression.

\paragraph{MLPs: Attention Knockout.} 
\label{6_1_2_mlps_attn_knockout}
Since MLPs themselves cannot attend to previous tokens--a function exclusive to MHA--we adopt an attention knockout approach inspired by \citet{geva2023dissecting}. By knocking out the attention from the last token to the target tokens, we examine changes in MLP output logits to determine how MLPs utilize target token information for knowledge recall and repetition avoidance. Specifically, we zero out the attention weights between the last and the target tokens:
\begin{equation*}
    p^{(l_i)}_{t_j} \gets 0, \forall i \in [1, n], \forall j \in [1, k], \forall l \in [1, L]
\end{equation*}

Let $m^{(l)}$ and $m'^{(l)}$ denote the MLP output at layer $l$ before and after applying the attention knockout respectively. We unembed these outputs using the same early decoding approach described in \Cref{4_1_method_decoding_components_outputs}. By subtracting the logits derived from $m'^{(l)}$ from those of $m^{(l)}$, we examine the difference in the logits of the subject and the answer tokens. A positive difference value indicates MLPs use the knocked-out tokens to promote a token; a negative difference means suppression. 

\subsection{Role of Subject Tokens}
\label{6_2_subject_tokens_for_knowledge_recall}

Across all models and datasets, attention and MLPs use subject tokens to contribute to answer promotions while suppressing the subject itself.

\begin{figure*}
    \centering
    \includegraphics[width=1\linewidth]{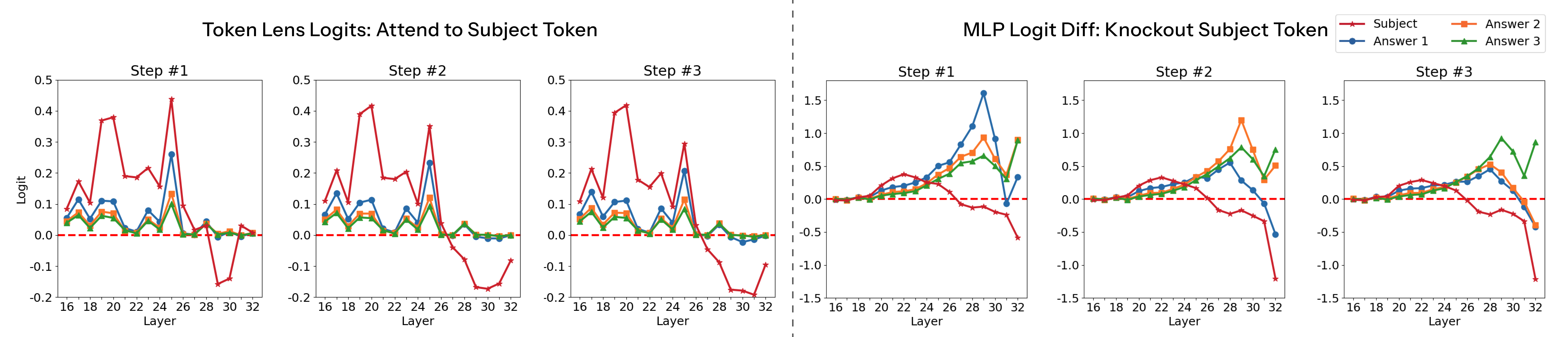}
    \caption{Token Lens logit values (left) and MLP logit differences (right) of subject and answer tokens when attending to or knocking out the subject tokens. Attention promotes and extracts subject information in the middle layers but suppresses it in later layers. MLPs promote the answers and suppress the subject at deeper layers. Refer to \Cref{appendix_critical_token_analysis_results} for full figures.
    }
    \label{fig:critical_token_analysis_subject_logits}
\end{figure*}

\paragraph{Attention first moves the subject to the last token position.} As shown in \Cref{fig:critical_token_analysis_subject_logits}, attention to the subject greatly increases the subject token's logit at the middle layers. To a lesser degree, it also promotes answer tokens, particularly at layer $25$.\footnote{Thus, the observation from \Cref{4_2_LMs_promote_then_suppress} that attention promotes answers at layer $25$ can be attributed to the subject token.} 
Answer promotion is most pronounced in the Country-Cities dataset (\Cref{fig:country_cities-llama-subject}, \Cref{fig:country_cities-mistral-subject}) but less evident in the Artist-Songs and Actor-Movies datasets (\Cref{fig:artist_songs-llama-subject}, \Cref{fig:artist_songs-mistral-subject}, \Cref{fig:actor_movies-llama-subject}, \Cref{fig:actor_movies-mistral-subject}). In all datasets, the subject logit is still larger than that of each answer across all answer steps, demonstrating that attention primarily copies or propagates subject information from the subject to the last token position.

\paragraph{MLPs use the subject to promote answers.} From the middle to late layers, MLP logit differences of the answer tokens are all positive across answer steps. Combined with attention's promotion of the subject, our findings suggest a coordinated mechanism: attention propagates subject information to the last token, and MLPs leverage this information to promote relevant answers. 

\paragraph{At late layers, attention shifts from promoting to suppressing the subject.} Starting around the $28$th layer, attention outputs negative logits for the subject tokens. %
This transition shows that while attention initially promotes the subject, it later suppresses the subject to prevent incorrect generations, as the subject itself is not a correct answer. 

\paragraph{MLPs amplify subject suppression.} The MLPs' logit differences for the subject token become negative in later layers, especially at steps $2$ and $3$. This pattern illustrates that MLPs not only promote answers but also actively suppress the subject when it is no longer relevant for the next prediction. Combined with attention's suppression of the subject at later stages, our result suggests that MLPs amplify suppression signals from attention to prevent incorrect generations.

\subsection{Role of Previous Answer Tokens}
\label{6_3_previous_answer_tokens_for_both_repetition_avoidance_and_knowledge_recall}

\begin{figure*}
    \centering
    \setlength{\abovecaptionskip}{1mm}
    \includegraphics[width=1\linewidth]{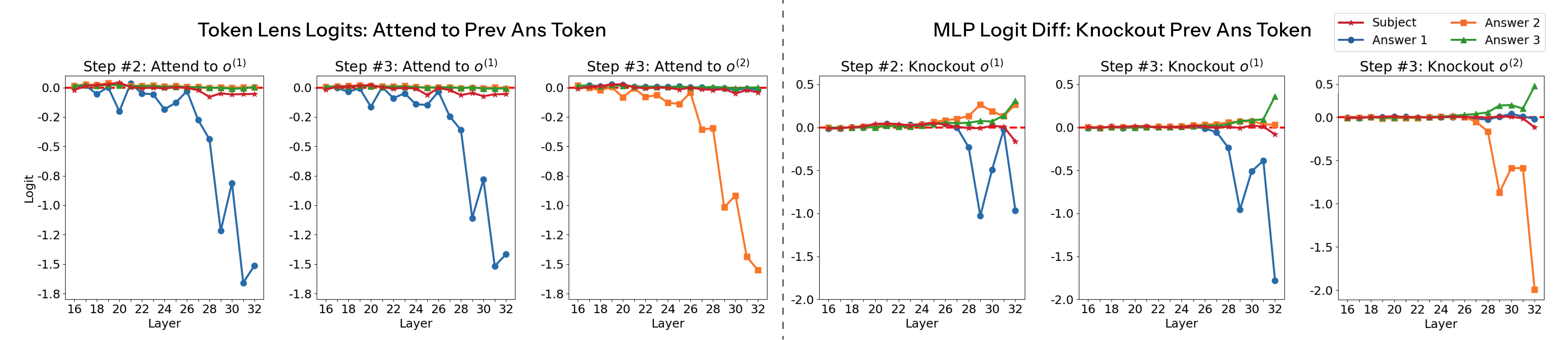}
    \caption{Token Lens logits of the attended previous answers (left) are negative at deeper layers, showing that attention suppresses prior answers. Negative MLP logit differences (right) for previous answers and positive differences for new answers suggest that MLPs use previous answer tokens for both repetition avoidance and knowledge recall.}
    \label{fig:critical_token_analysis_prev_ans_logits}
    \vspace{-3mm}
\end{figure*}

\paragraph{Attention plays a crucial role in suppressing repetitions.} Attention consistently outputs negative logits for previous answer tokens at both step $2$ and step $3$ in the final layers. This result shows that attention attends to and suppresses tokens that have already appeared in the context, ensuring previously generated answers are not repeated. 

\paragraph{MLPs amplify suppression of previous answers.} As shown in \Cref{fig:critical_token_analysis_prev_ans_logits}, all previous answer tokens have negative MLP logit differences at late layers. For instance, $o_1$ has negative logits at step $2$ starting around layer $27$; $o_1$ and $o_2$ exhibit similar patterns at step $3$. This suppression aligns with attention’s role in inhibiting previously generated tokens, suggesting that MLPs amplify these suppression signals to prevent repetition.

\paragraph{MLPs also use previous answer tokens for knowledge recall.} Surprisingly, we observe positive MLP logit differences for new answers across answer steps (\Cref{fig:critical_token_analysis_prev_ans_logits}). Specifically, the logit differences of both $o_2$ and $o_3$ are positive when intervening on $o_1$ at step $2$; $o_3$ has positive logits differences when intervening on $o_1$ or $o_2$ at step $3$. This pattern shows that MLPs also leverage previous answer tokens to promote new answers. Since LMs already promote all relevant answers when predicting previous answers, it is plausible that the models reuse these prior computations to promote new answers.
These findings show that the subject token is not the sole source of answer promotion (\Cref{5_1_which_tokens_should_be_noised}). The previous answer tokens also have a positive (but smaller) effect on answer promotion.

\subsection{Role of Last Token}

The last token aggregates knowledge recall and suppression information in the final layers to promote all answers while prioritizing the correct answer for each step.

\begin{figure*}
    \centering
    \setlength{\abovecaptionskip}{1mm}
    \includegraphics[width=1\linewidth]{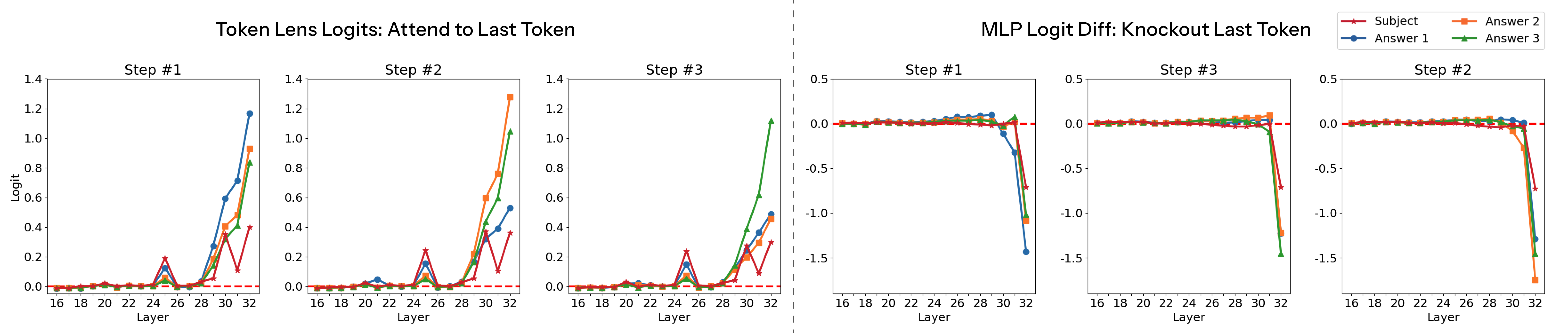}
    \caption{
    Token Lens logit values (left) and MLP logit differences (right) of subject and answer tokens when attending to or knocking out the last token. Attention promotes all three answers and the subject at the final layers, prioritizing $o^{(i)}$ at each step $i$. Late-layer MLP logit differences are negative for the subject and answers, possibly compensating for the absence of direct attention to the last token to encourage correct outputs.
    }
    \label{fig:critical_token_analysis_last_token_logits}
    \vspace{-3mm}
\end{figure*}

\paragraph{Attention promotes all answers at the last token in the final layers.} Starting from layer $28$, attention from the last token to itself significantly increases the logit of all three answers (\Cref{fig:critical_token_analysis_last_token_logits}). At each step $i$, the logit for the answer ${o^{(i)}}$ is consistently the highest among the three answers. This result suggests that attention at the last token aggregates information from earlier layers related to knowledge recall and suppression, preparing the model for generating the next prediction. 

\paragraph{MLPs compensate answer promotions when the direct attention to the last token is absent.} Interestingly, we observe MLPs output negative logit differences for the subject and all three answers in the final layers when knocking out the attention from the last token to itself (\Cref{fig:critical_token_analysis_last_token_logits}). The answer ${o^{(i)}}$ for each step $i$ consistently has the most negative logit differences. In other words, without having access to the attention output of the last token, MLPs output even higher logits for the subject and the answers. This behavior suggests a backup mechanism: without direct attention to the last token that aggregates information from early input tokens, the model may not have sufficient promotion and differentiation of the three answers. MLPs compensate this by further promoting the three answers to encourage the predictions to be correct. Similarly, \citet{wang2022interpretability} find backup token mover attention heads that become active when the original token mover heads are ablated. %

\section{Are Knowledge Recall and Suppression Independent?}
Observing that LMs promote all answers while suppressing previously generated ones, another question we have is whether knowledge recall and suppression are independent. To investigate this, we analyze the behavior of individual attention heads at the last token position to determine if they perform one, both, or neither of the two subtasks.

\begin{figure*}[!]
    \centering
    \includegraphics[width=1\linewidth]{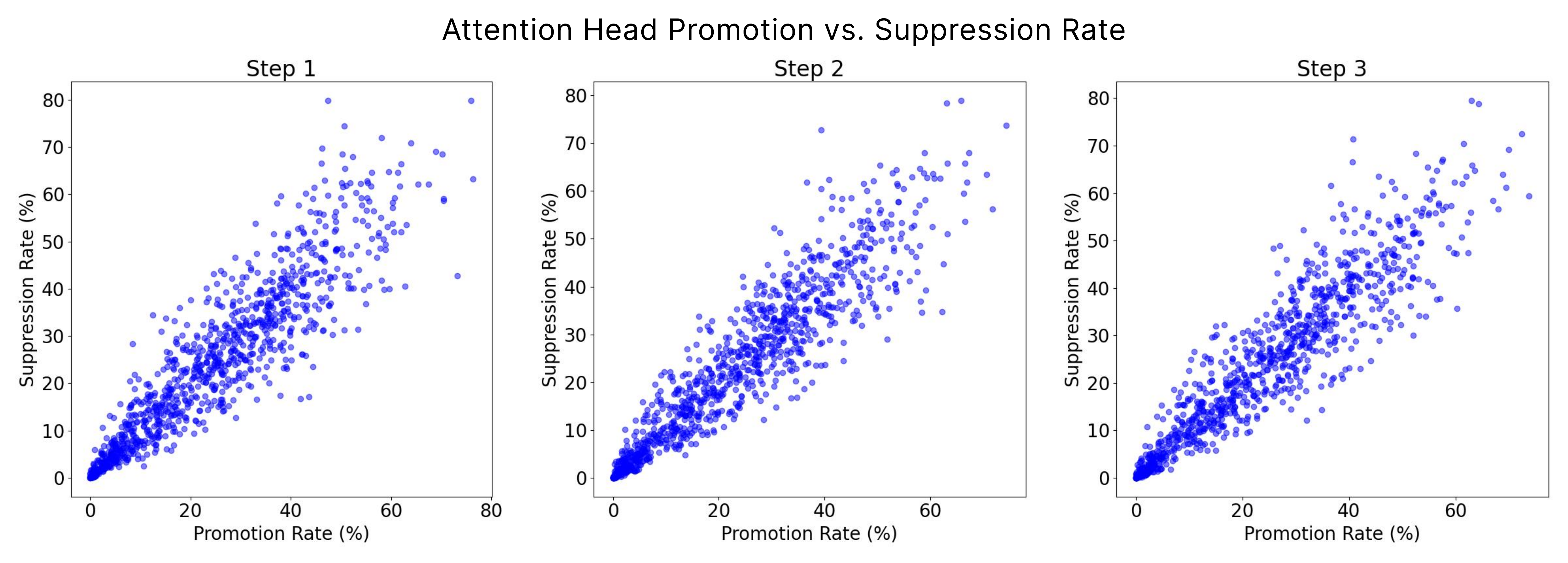}
    \caption{Promotion rate versus suppression rate of all attention heads across three answer steps macro-averaged across all models and datasets with template $1$. The promotion rate and suppression rate positively correlate with each other, suggesting that answer promotion and suppression may not be independent of each other.}
    \label{fig:head_promotion_vs_suppression_rate}
\vspace{-5mm}
\end{figure*}

\subsection{Methodology: Characterizing Attention Heads' Behavior}
Our methodology involves the following steps:
\begin{enumerate}
    \item Decode Attention Head Outputs: For each attention head, we decode its output at the last token position and collect the logits of the first token of $t$ for a given input, where $t \in \{s, o^{(1)}, o^{(2)}, o^{(3)}\}$.
    
    \item Calculate Layer-wise Baseline: For each layer $l$, we compute the mean $\mu_l$ and standard deviation $\sigma_l$ of attention head logits across all heads in the layer.
    
    \item Characterize Head Behavior: Let $\text{logit}(a^{(li)}_{t})$ denote the logit for the first token of $t$ from attention head $a^{(li)}$. The behavior of $a^{(li)}$ on token $t$ is classified as:\\
    
    \scalebox{0.75}{$
        \text{Behavior}(a^{(li)}_{t})=
                \begin{cases}
                    \text{Promotion},   & \text{if logit}(a^{(li)}_t) > \mu_l + \sigma_l \\
                    \text{Suppression}, & \text{if logit}(a^{(li)}_t) < \mu_l - \sigma_l \\
                    \text{None},        & \text{otherwise}
                \end{cases}
    $}
    
    \item Classify Head Function: $a^{(li)}$ is classified as performing promotion for the given input if it promotes the first token of any $t \in {s, o^{(1)}, o^{(2)}, o^{(3)}}$ and as performing suppression if it suppresses any such token.

    \item Aggregate Results: We average the percentage of times each attention head is identified as performing promotion or suppression.
\end{enumerate}

Then, by plotting the promotion rate against the suppression rate for all heads, we examine how LMs divide the labor among heads for knowledge recall and suppression.

\subsection{Knowledge Recall and Suppression May Not be Independent}

As can be observed in \Cref{fig:head_promotion_vs_suppression_rate}, the promotion rate and suppression rate of attention heads consistently correlate with each other across all three answers steps, with the majority of the data points concentrated in the bottom-left region of the plots. This finding shows that most attention heads contribute moderately to the two subtasks and are responsible for both token promotion and suppression, suggesting that knowledge recall and suppression may not be independent. 
\section{Conclusion}

We uncover how language models answer one-to-many factual queries across two models and three datasets. By unembedding the output of attention and MLPs across layers, we find that LMs promote all answers and then suppress previously generated ones. We then delve into how LMs implement knowledge recall and repetition avoidance. We find that LMs use both the subject and previous answer tokens to perform knowledge recall. Attention first propagates subject information from the subject to the last token, which is then used by MLPs to promote all correct answers. At the same time, MLPs also utilize previous answer tokens to promote new answers at late layers. %
In addition, previous answer tokens trigger suppression of themselves. In the final layers, attention suppresses repetitions by attending to and outputting negative logits for previously generated answer tokens. MLPs reinforce and amplify this suppression by decreasing the logits of previous answer tokens around the same layers. At last, by integrating all relevant information for knowledge recall and suppression at the last token position, LMs effectively generate correct and distinct answers at different steps. We hope our findings encourage a deeper understanding of how LMs' internal components interact with context tokens to support complex factual recall and response generation.

\paragraph{Future Work.} Future work could investigate possible redundancies in the model, as multiple tokens---such as the subject and previous answers---contribute to promoting new answers. This result raises the question of whether LMs redundantly encode knowledge and if it is necessary. Additionally, our analyses only focus on the correct cases. Examining the patterns when LMs use unreliable signals for factual recall or hallucinate could provide insights for mitigating such errors \cite{saynova2024fact}.

\section*{Limitations}
Our analyses primarily rely on Logit Lens \cite{Nostalgebraist_2020}, which early decodes component outputs using LMs' last unembedding layer. While this method is training-free, it may be less reliable, particularly for early layers. More expressive techniques, such as Tuned Lens \cite{belrose2023eliciting} and SAE \cite{templeton2024scaling}, could be applied for a better understanding of \task. Also, we use a single prompt template for each model and dataset. Further studies are needed to determine whether our findings generalize across different prompt templates. 

While we attempt to identify how LMs recall knowledge, it is difficult to disentangle where the model truly recalls knowledge from its parameters, and where it amplifies already-recalled knowledge stored in the residual stream.
This is especially difficult because models could redundantly encode knowledge in multiple places, and thus parametric recall and amplification could be interleaved.
We hope future work can develop reliable methods for disentangling these concepts and lead to a more precise understanding of the underlying mechanism.

\section*{Acknowledgments}
We thank Ting-Yun Chang for her helpful feedback on the paper. RJ was supported in part by the National Science Foundation under Grant No. IIS-2403436. Any opinions, findings, and conclusions or recommendations expressed in this material are those of the author(s) and do not necessarily reflect the views of the National Science Foundation.

\bibliography{reference}
\bibliographystyle{acl_natbib}
\clearpage
\appendix 
\section{Prompt Templates}
\label{appendix_prompts}
Refer to \Cref{tab:prompt-templates} for the prompt templates that we use for each model and dataset. See \Cref{tab:model_performance_on_three_templates} for number of correct cases from each model, dataset, and template.
\renewcommand{\arraystretch}{1.3}
\begin{table*}[ht]
\centering
\footnotesize
\begin{tabular}{c|c|p{3.5cm}p{3.8cm}p{3.5cm}}
\toprule
\textbf{Dataset} & \textbf{Model} & \multicolumn{1}{c}{\textbf{Template 1}} & \multicolumn{1}{c}{\textbf{Template 2}} & \multicolumn{1}{c}{\textbf{Template 3}} \\
\hline
\multirow[c]{2}{*}{Country-Cities} 
    & Llama & List three cities from \textless country\textgreater & Name three cities located in \textless country\textgreater & Give the names of three cities in \textless country\textgreater \\
    \cline{2-5}
    & Mistral & List the name of three cities from \textless country\textgreater & Provide just the names of three cities in \textless country\textgreater & State three city names from \textless country\textgreater \\
\hline
\multirow[c]{2}{*}{Artist-Songs} 
    & Llama & List three songs performed by \textless artist\textgreater & Name three songs sung by \textless artist\textgreater & Mention three tracks performed by \textless artist\textgreater \\
    \cline{2-5}
    & Mistral & List three songs performed by \textless artist\textgreater & Provide just the names of three songs by \textless artist\textgreater & State three song titles by \textless artist\textgreater \\
\hline
\multirow[c]{2}{*}{Actor-Movies} 
    & Llama & List three movies acted by actor \textless actor\textgreater & Name three movies that feature \textless actor\textgreater & Mention three films that include \textless actor\textgreater \\
    \cline{2-5}
    & Mistral & List the name of three movies acted by \textless actor\textgreater & Provide just the names of three movies featuring \textless actor\textgreater & State three movie titles starring \textless actor\textgreater \\
\bottomrule
\end{tabular}
\caption{Prompt templates used across datasets and models. Llama is short for Llama-3-8B-Instruct. Mistral is short for Mistral-7B-Instruct-v0.2.}
\label{tab:prompt-templates}
\end{table*}
\renewcommand{\arraystretch}{1.3}
\begin{table*}[ht]
\centering
\small
\begin{tabular}{c|c|c|c|c}
\toprule
\textbf{Dataset} & \textbf{Model} & \textbf{Template 1} & \textbf{Template 2} & \textbf{Template 3} \\
\hline
\multirow{2}{*}{Country-Cities} 
    & Llama-3-8B-Instruct       & 122/168 (72.6\%)   & 122/168 (72.62\%)  & 123/168 (73.21\%) \\
    & Mistral-7B-Instruct-v0.2 & 116/168 (69.0\%)   & 122/168 (72.62\%)  & 116/168 (69.05\%) \\
\hline
\multirow{2}{*}{Artist-Songs} 
    & Llama-3-8B-Instruct       & 261/2077 (12.6\%)  & 287/2077 (13.82\%) & 279/2077 (13.43\%) \\
    & Mistral-7B-Instruct-v0.2 & 206/2077 (9.9\%)   & 240/2077 (11.56\%) & 217/2077 (10.45\%) \\
\hline
\multirow{2}{*}{Actor-Movies} 
    & Llama-3-8B-Instruct       & 1285/8790 (14.6\%) & 1263/8790 (14.37\%) & 1157/8790 (13.16\%) \\
    & Mistral-7B-Instruct-v0.2 & 965/8790 (11.0\%)  & 905/8790 (10.30\%)  & 528/8790 (6.01\%) \\
\bottomrule
\end{tabular}
\caption{Number of correct cases and accuracy of two models on each dataset and template.}
\label{tab:model_performance_on_three_templates}
\end{table*}

\section{Sample Responses and Example of Analysis Data Creation}
\label{appendix_sample_responses_and_analysis_data_example}
Following are some sample responses from the model:
Llama-3-8B-Instruct:
\begin{itemize}
    \item List three cities from China: 1. Beijing 2. Shanghai 3. Guangzhou
    \item List three songs performed by Ed Sheeran: 1. "Shape of You" 2. "Thinking Out Loud" 3. "Photograph"
    \item List three movies acted by Meryl Streep: 1. The Devil Wears Prada (2006) 2. The Iron Lady (2011) 3. Sophie's Choice
\end{itemize}

Mistral-7B-Instruct-v0.2:
\begin{itemize}
    \item List the name of three cities from China:\textbackslash n\textbackslash n1. Beijing\textbackslash n2. Shanghai\textbackslash n3. Guangzhou
    \item List the name of three songs performed by Ed Sheeran:  1. Shape of You, 2. Perfect, 3. Thinking Out Loud
    \item List the name of three movies acted by Meryl Streep:  1. The Devil Wears Prada (2006)\textbackslash n2. Sophie's Choice (1982)\textbackslash n3. Kramer vs. Kramer
\end{itemize}

We filter out responses that contain incorrect object entities and only focus on the correct cases for analyses. For the artist-songs dataset, we use the Spotify API \footnote{\url{https://developer.spotify.com/documentation/web-api}}  to extend the song lists and keep them more up-to-date. To create data for analyzing, for example, Mistral-7B-Instruct-v0.2's behavior when predicting the first answer about Ed Sheeran, we will use "List the name of three songs performed by Ed Sheeran:  1." as the input and examine models' behavior when predicting "Shape".

\section{Decoding Attention and MLP Outputs Results}
\label{appendix_unembed_attn_mlp_outputs_full_figures}
\begin{figure*}
    \centering
    \includegraphics[width=1\linewidth]{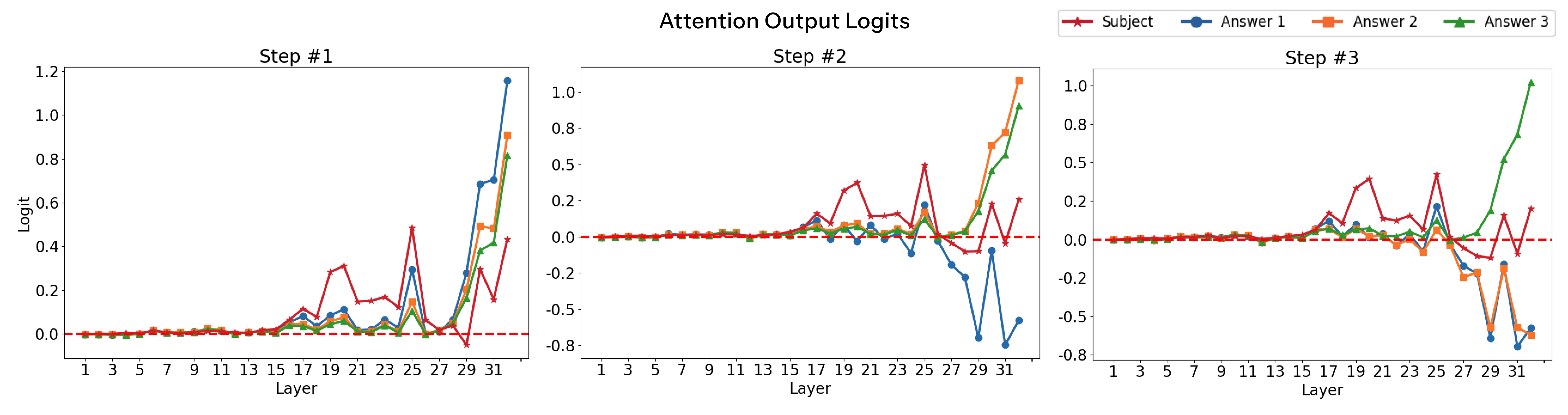}
    \caption{Logit of the subject and answer tokens from decoding the attention outputs across layers and answer steps. Attention primarily promotes the subject at the middle layers while promoting new answers and suppressing previously generated ones at deeper layers.}
    \label{fig:unembed_attn_output_logit}
\end{figure*}

\begin{figure*}
    \centering
    \includegraphics[width=1\linewidth]{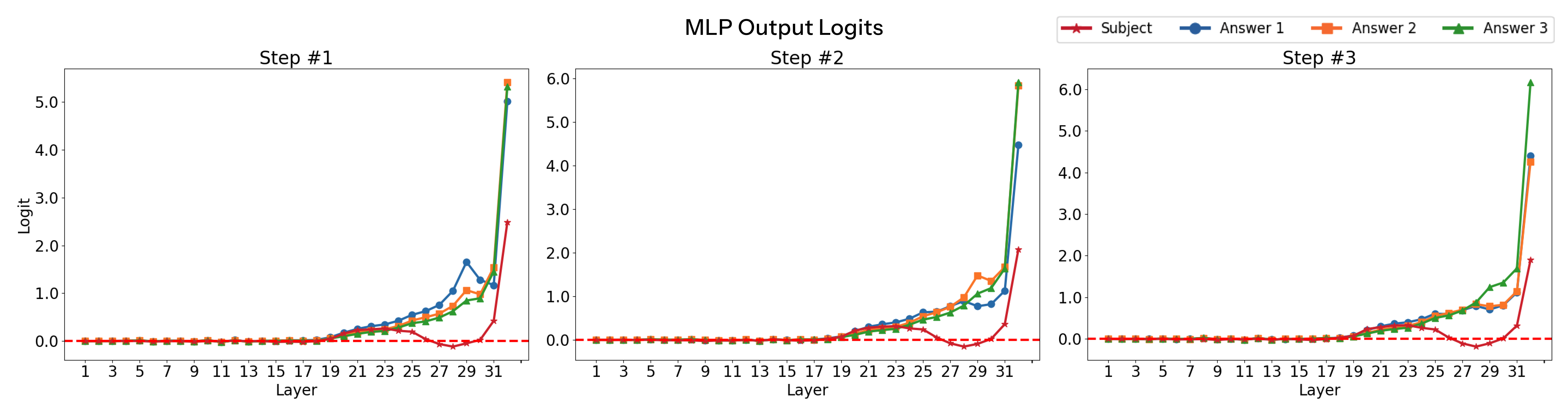}
    \caption{Logits of the subject and answer tokens from decoding the MLP outputs across layers and answer steps. The consistently positive logits for all three answers illustrate that MLPs promote multiple answers simultaneously. MLPs also decrease the logits of previously generated answers in deeper layers, contributing to repetition suppression alongside attention.
    }
    \label{fig:unembed_mlp_output_logit}
\end{figure*}

\begin{figure*}
   \centering
   \includegraphics[width=1\linewidth]{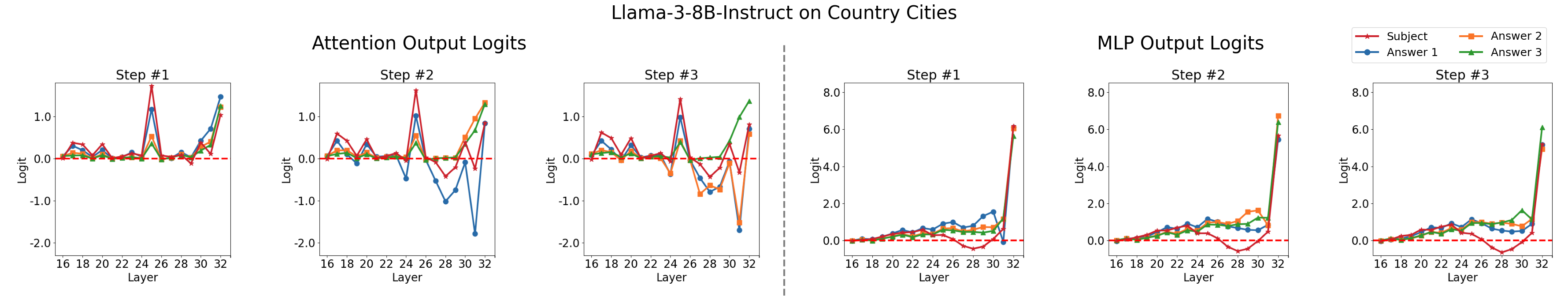}
   \caption{Attention and MLP output logits of Llama-3-8B-Instruct on Country-Cities dataset averaged across three prompt templates.}
   \label{fig:country_cities-llama-attn_mlp_output_logit}
\end{figure*}

\begin{figure*}
   \centering
   \includegraphics[width=1\linewidth]{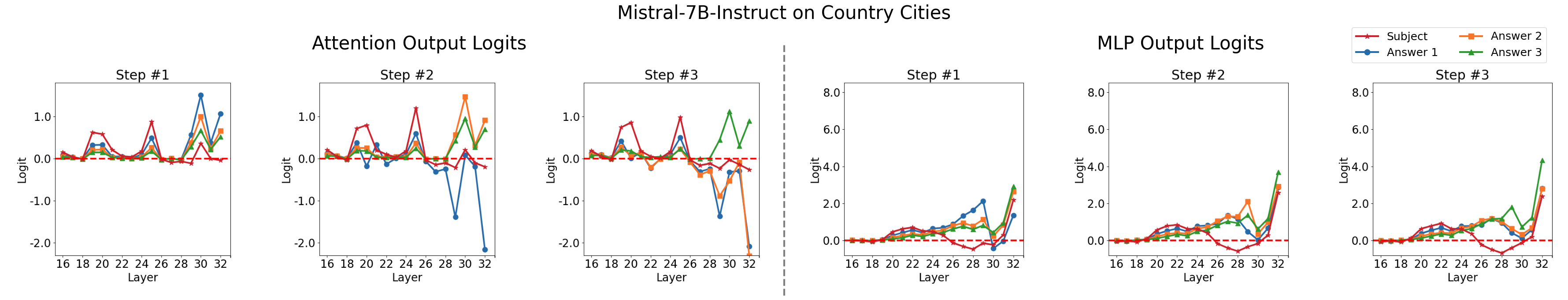}
   \caption{Attention and MLP output logits of Mistral-7B-Instruct on Country-Cities dataset averaged across three prompt templates.}
   \label{fig:country_cities-mistral-attn_mlp_output_logit}
\end{figure*}

\begin{figure*}
   \centering
   \includegraphics[width=1\linewidth]{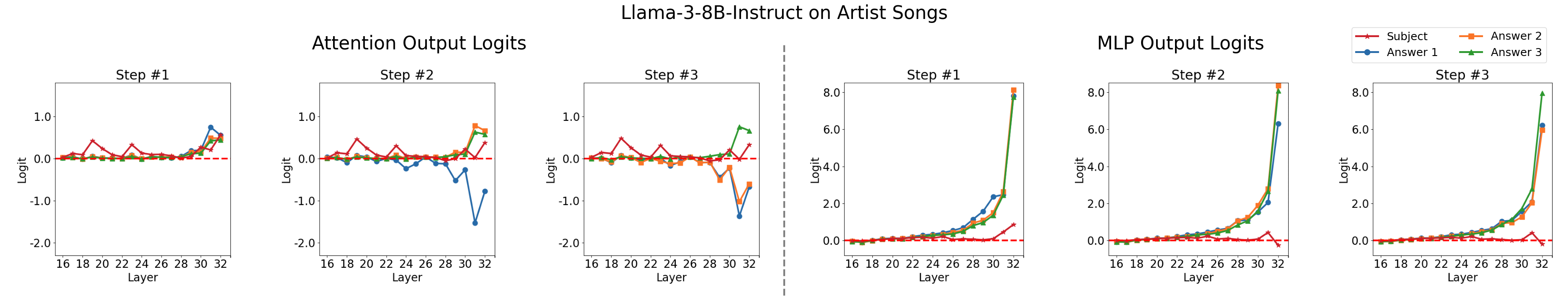}
   \caption{Attention and MLP output logits of Llama-3-8B-Instruct on Artist-Songs dataset averaged across three prompt templates.}
   \label{fig:artist_songs-llama-attn_mlp_output_logit}
\end{figure*}

\begin{figure*}
   \centering
   \includegraphics[width=1\linewidth]{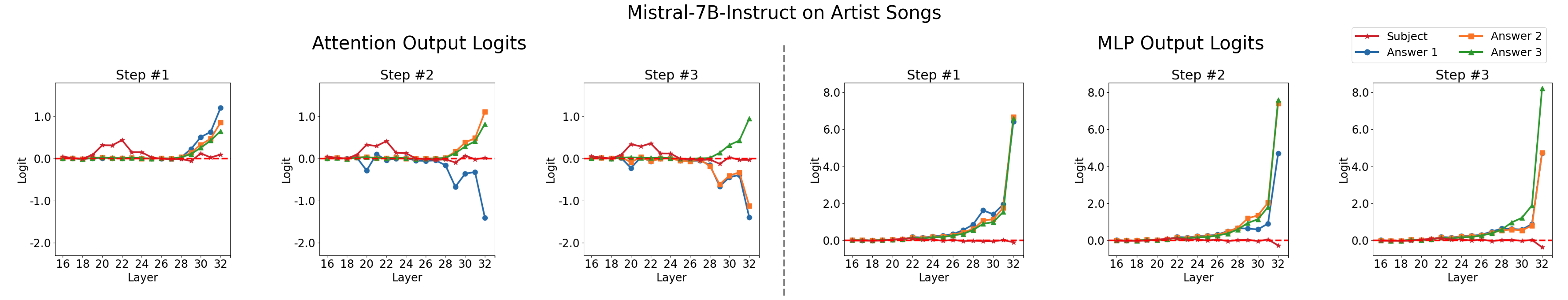}
   \caption{Attention and MLP output logits of Mistral-7B-Instruct on Artist-Songs dataset averaged across three prompt templates.}
   \label{fig:artist_songs-mistral-attn_mlp_output_logit}
\end{figure*}

\begin{figure*}
   \centering
   \includegraphics[width=1\linewidth]{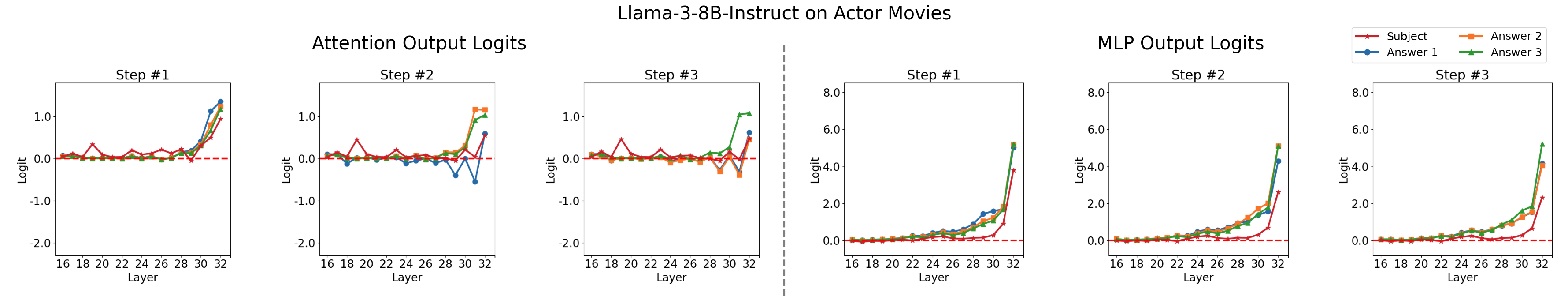}
   \caption{Attention and MLP output logits of Llama-3-8B-Instruct on Actor-Movies dataset averaged across three prompt templates.}
   \label{fig:actor_movies-llama-attn_mlp_output_logit}
\end{figure*}

\begin{figure*}
   \centering
   \includegraphics[width=1\linewidth]{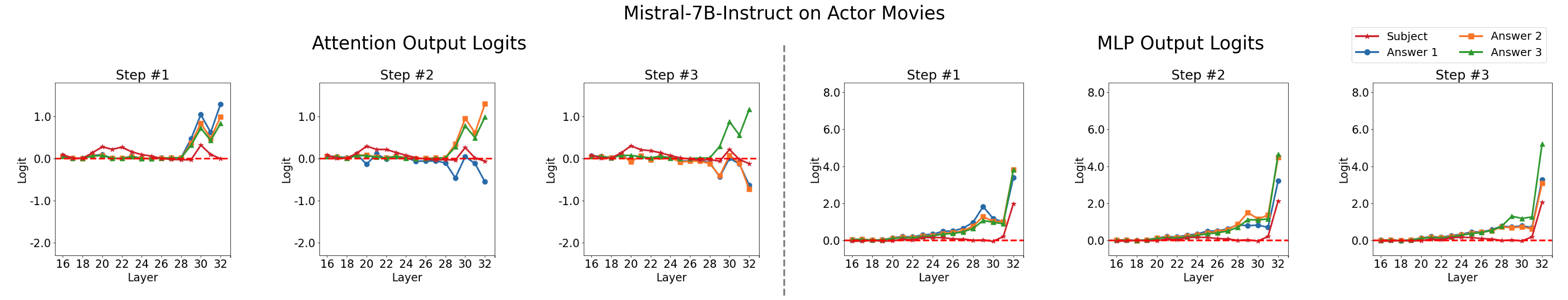}
   \caption{Attention and MLP output logits of Mistral-7B-Instruct on Actor-Movies dataset averaged across three prompt templates.}
   \label{fig:actor_movies-mistral-attn_mlp_output_logit}
\end{figure*}

\Cref{fig:unembed_attn_output_logit} and \Cref{fig:unembed_mlp_output_logit} are the full figures of logit of the subject and target entity tokens from decoding attention and mlp output across layers and answer steps. \Cref{fig:country_cities-llama-attn_mlp_output_logit}, \Cref{fig:country_cities-mistral-attn_mlp_output_logit}, \Cref{fig:artist_songs-llama-attn_mlp_output_logit}, \Cref{fig:artist_songs-mistral-attn_mlp_output_logit}, \Cref{fig:actor_movies-llama-attn_mlp_output_logit}, \Cref{fig:actor_movies-mistral-attn_mlp_output_logit} are the figures for specific models and datasets. As can be seen from \Cref{fig:country_cities-llama-attn_mlp_output_logit} and \Cref{fig:country_cities-mistral-attn_mlp_output_logit}, attention performing answer promotion at middle layers is more evident in the Country-Cities dataset. However, it is much less evident in the other two datasets (\Cref{fig:artist_songs-llama-attn_mlp_output_logit}, \Cref{fig:artist_songs-mistral-attn_mlp_output_logit}, \Cref{fig:actor_movies-llama-attn_mlp_output_logit}, \Cref{fig:actor_movies-mistral-attn_mlp_output_logit}). Refer to \url{https://drive.google.com/drive/folders/1Xnk3lPLuqjmNABfrJvcJ4mSM9EBvYoub?dmr=1&ec=wgc-drive-globalnav-goto} for the figures without early layers omitted.

\section{Causal Tracing Results}
\label{appendix_causal_tracing_figures}
\begin{figure*}
    \centering
    \includegraphics[width=1\linewidth]{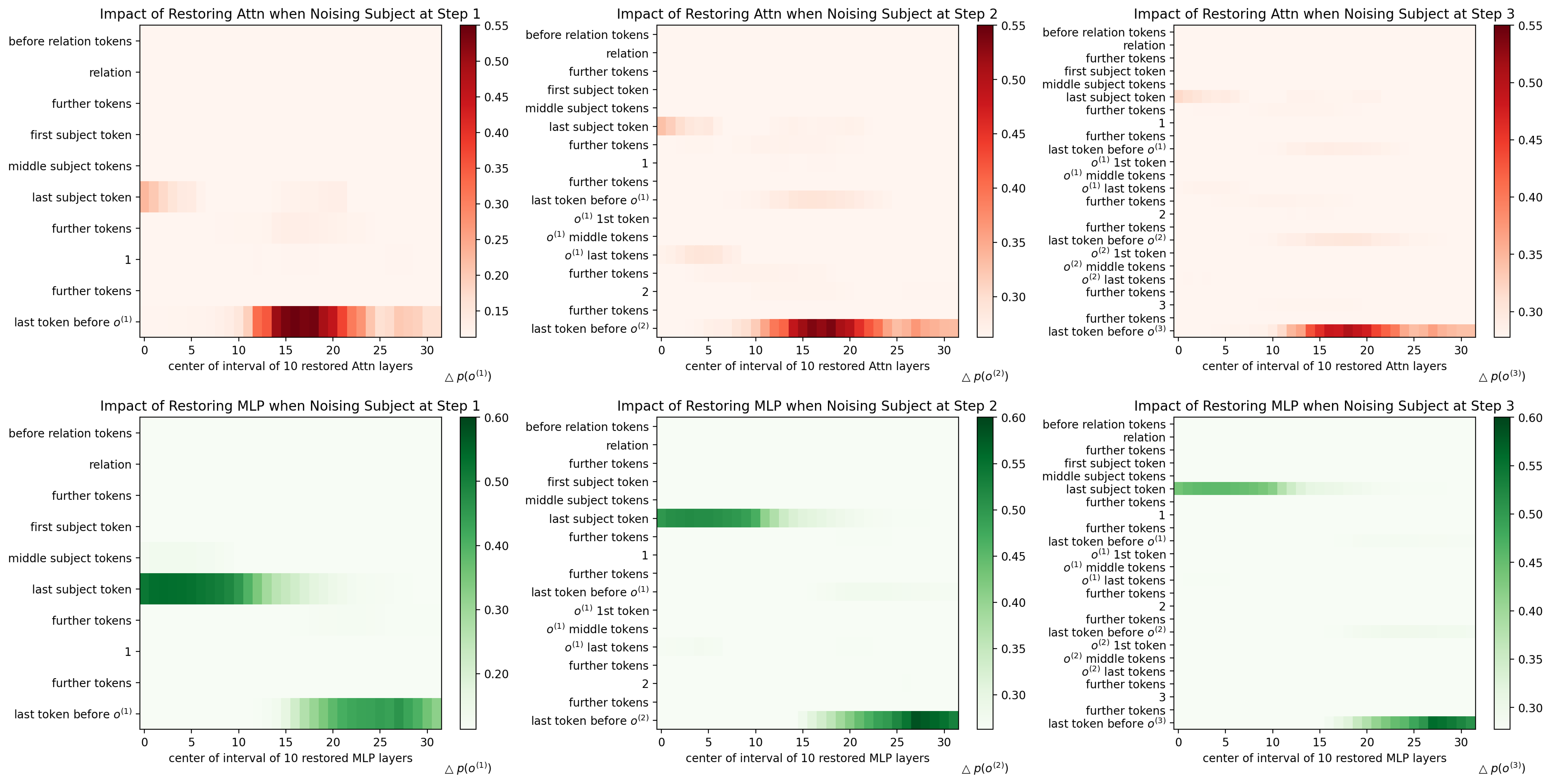}
    \caption{The impact of attention and MLPs' activations on LMs' predictions when intervening on the subject tokens across three answer steps macro-averaged across all models, templates, and 100 instances per dataset. Attention contributions dominate in the middle layers at the last token, while MLPs are important in early layers at the subject token and in late layers at the last token. The probability differences all peak around or above $0.55$, reflecting the importance of the subject tokens. 
    }
    \label{fig:subject_causal_tracing}
\end{figure*}

\begin{figure*}
    \centering
    \includegraphics[width=1\linewidth]{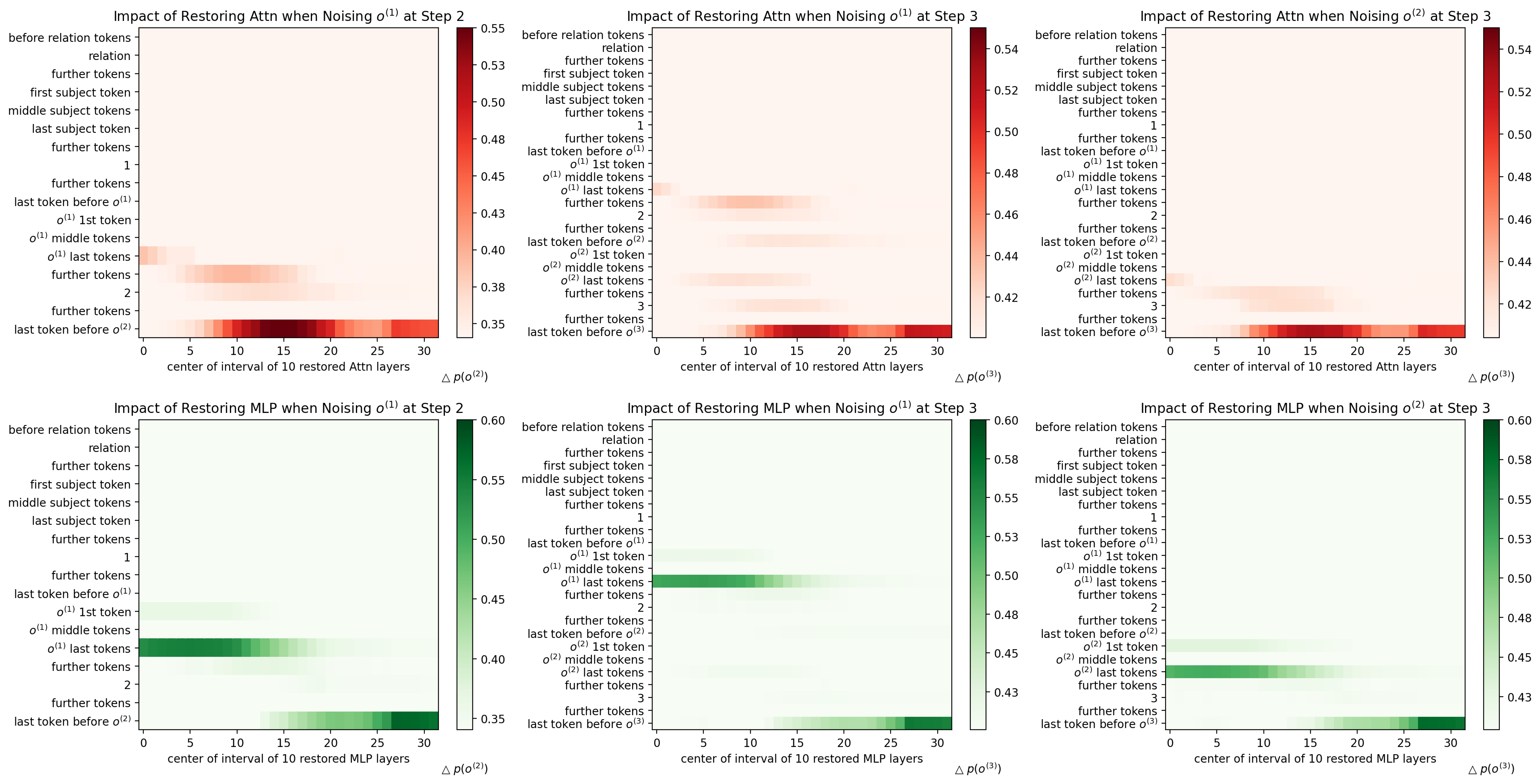}
    \caption{The impact of attention and MLPs' activations on LMs' predictions when intervening on previous answer tokens at step $2$ and $3$ macro-averaged across all models, templates, and 100 instances per dataset. Attention is important in both the middle and the last layers at the last token position. MLPs' contributions are critical in early layers at the previous answer positions and in final layers at the last token. The probability differences all peak around or above $0.54$, indicating previous answer tokens are critical to models' predictions.}
    \label{fig:prev_ans_causal_tracing}
\end{figure*}

\Cref{fig:subject_causal_tracing} and \Cref{fig:prev_ans_causal_tracing} are the full figures for causal tracing when noising the subject and previous answer tokens across all three answer steps and templates. Refer to \url{https://drive.google.com/drive/folders/1aG-GZEIZ_EgUKQ8Vhe_Lv0mHILxIZfms?dmr=1&ec=wgc-drive-globalnav-goto} for figures of specific models and datasets.


\section{Critical Token Analysis Results}
\label{appendix_critical_token_analysis_results}
\begin{figure*}
    \centering
    \includegraphics[width=1\linewidth]{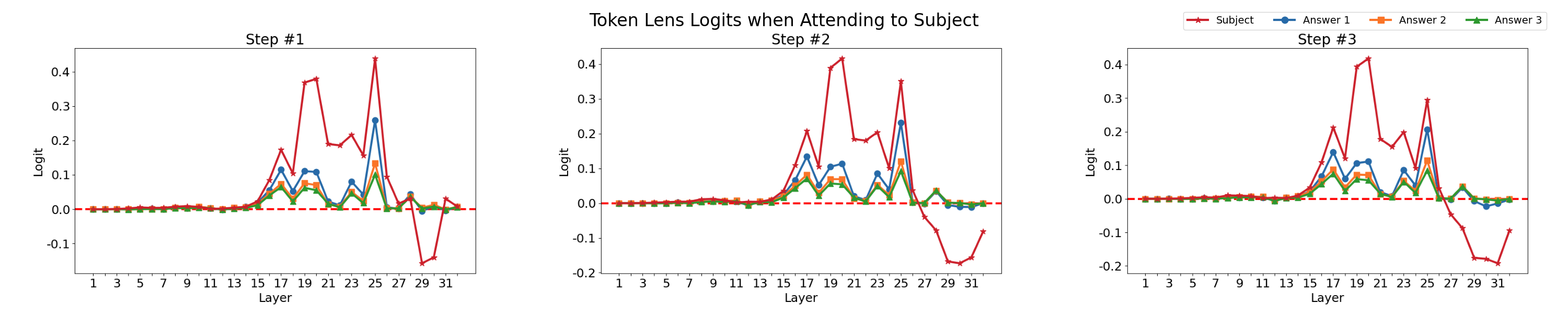}
    \caption{Token Lens logit values of subject and answer tokens across layers and answer steps when attending to the subject (macro-averaged across all datasets, models, and templates). Attention promotes and extracts subject information in the middle layers while suppressing it in later layers.}
    \label{fig:token_lens_logit_attention_to_subject}
\end{figure*}

\begin{figure*}
    \centering
    \includegraphics[width=1\linewidth]{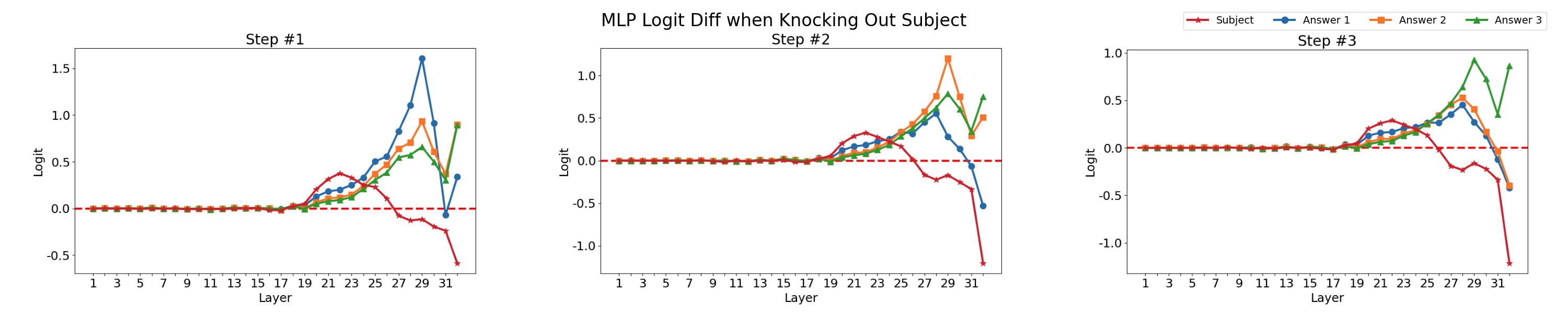}
    \caption{Logit differences of the subject and answer tokens between MLP outputs with and without knocking out attention from the last to the subject tokens (macro-averaged across all datasets, models, and templates). Positive logit differences for the answers and negative differences for the subject in later layers show that MLPs use the subject information to promote answers and suppress the subject.}
    \label{fig:attention_knockout_subject_logits}
\end{figure*}

\begin{figure*}
    \centering
    \includegraphics[width=1\linewidth]{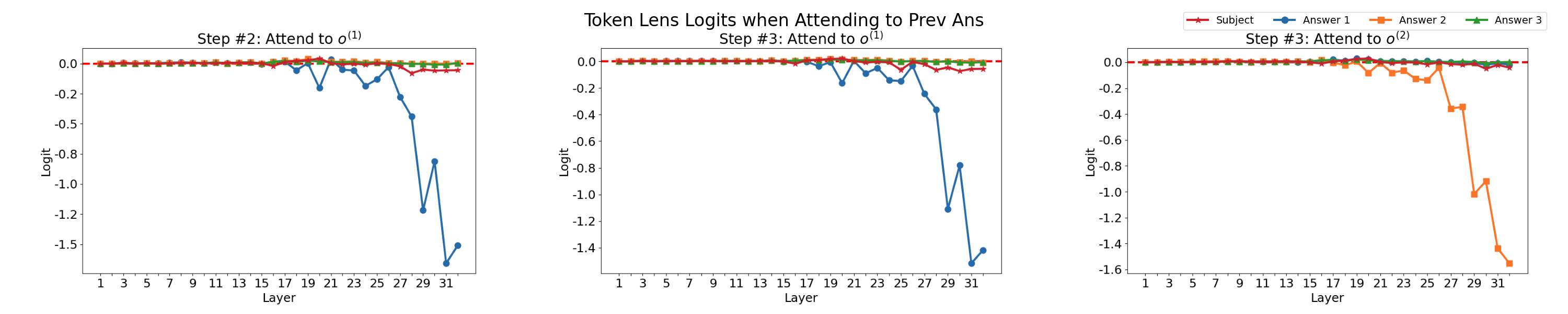}
    \caption{Token Lens logit values subject and answer tokens across layers and answer steps $2$ and $3$ (macro-averaged across all datasets, models, and templates) when attending to previous answers. The logit of the attended answer is negative at later layers, showing that the attention is suppressing previously generated answers.}
    \label{fig:token_lens_logit_attention_to_prev_ans}
\end{figure*}

\begin{figure*}
    \centering
    \includegraphics[width=1\linewidth]{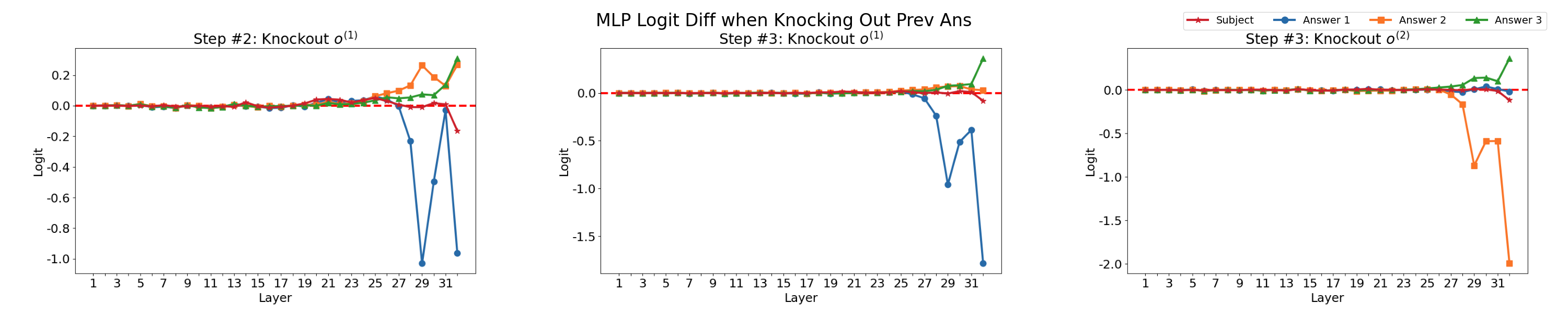}
    \caption{Logit differences for subject and answer tokens between MLP outputs with and without knocking attention from the last to previous answer tokens (macro-averaged across all datasets, models, and templates). All previously generated answer tokens have negative logits, and all new answers have positive logits. This result suggests that MLPs use previous answers for both repetition suppression and new answer promotion.}
    \label{fig:attention_knockout_prev_ans_logits}
\end{figure*}

\begin{figure*}
    \centering
    \includegraphics[width=1\linewidth]{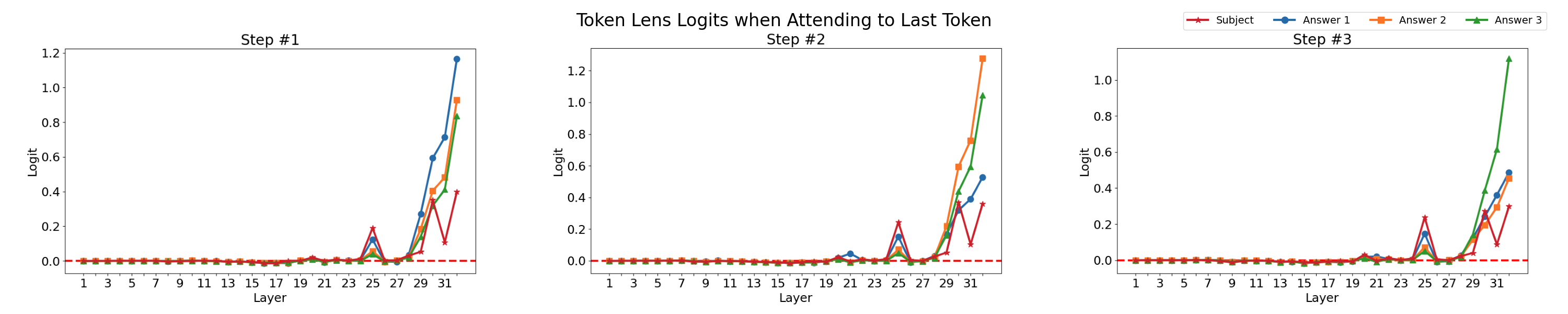}
    \caption{Token Lens logit values of subject and answer tokens across layers and answer steps when attending to the last token (macro-averaged across all datasets, models, and templates). Attention promotes all three answers and the subject at the final layers, with the answer for the current step having the highest logit.}
    \label{fig:token_lens_last_token_logits}
\end{figure*}

\begin{figure*}
    \centering
    \includegraphics[width=1\linewidth]{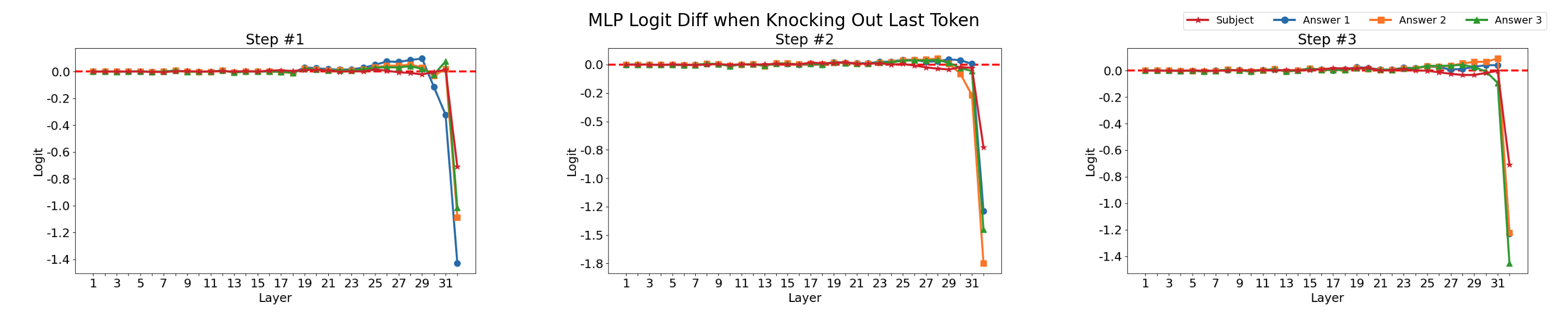}
    \caption{Logit differences for subject and answer tokens between MLP outputs with and without knocking attention from the last token to itself (macro-averaged across all datasets, models, and templates). The logit differences of all three answers and the subject are negative at the late layers, meaning MLPs output higher logits when it does not have information from the last token. This pattern may suggest a compensation behavior for the absence of direct attention to the last token to encourage the outputs to still be correct.}
    \label{fig:attention_knockout_last_token_logits}
\end{figure*}

\begin{figure*}
   \centering
   \includegraphics[width=1\linewidth]{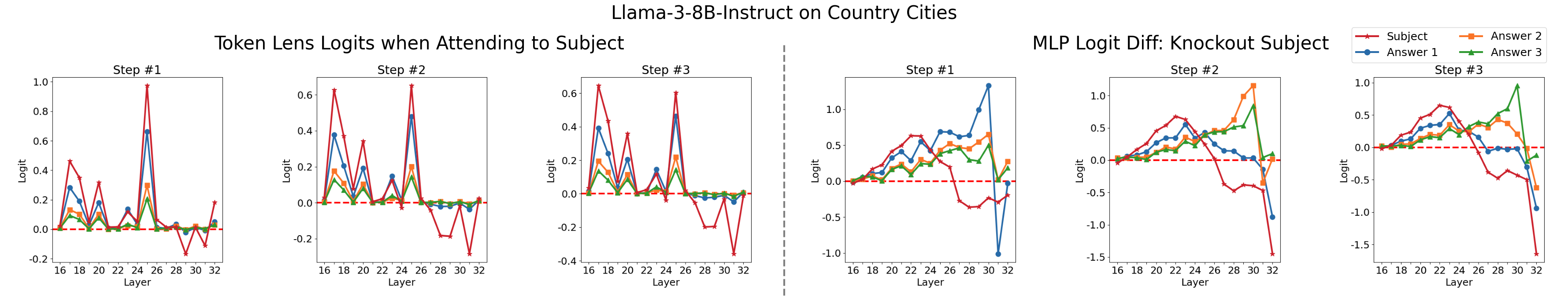}
   \caption{Token Lens logit values (left) and MLP logit differences (right) of subject and answer tokens of Llama-3-8B-Instruct on Country-Cities dataset when attending to or knocking out the subject tokens (averaged across three prompt templates).}
   \label{fig:country_cities-llama-subject}
\end{figure*}

\begin{figure*}
   \centering
   \includegraphics[width=1\linewidth]{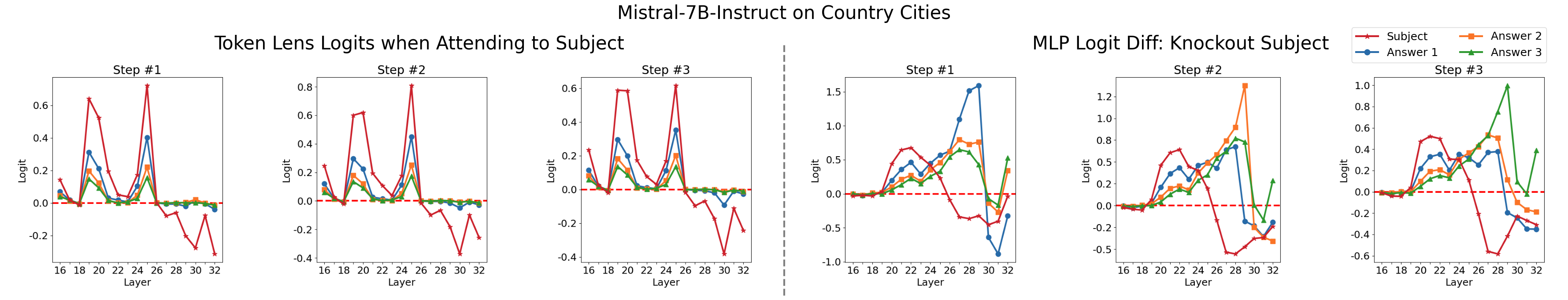}
   \caption{Token Lens logit values (left) and MLP logit differences (right) of subject and answer tokens of Mistral-7B-Instruct on Country-Cities dataset when attending to or knocking out the subject tokens (averaged across three prompt templates).}
   \label{fig:country_cities-mistral-subject}
\end{figure*}

\begin{figure*}
   \centering
   \includegraphics[width=1\linewidth]{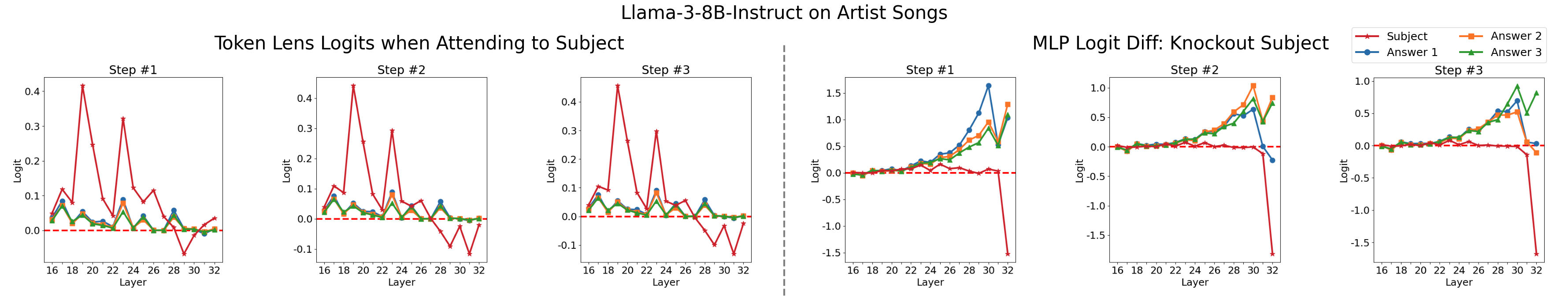}
   \caption{Token Lens logit values (left) and MLP logit differences (right) of subject and answer tokens of Llama-3-8B-Instruct on Artist-Songs dataset when attending to or knocking out the subject tokens (averaged across three prompt templates).}
   \label{fig:artist_songs-llama-subject}
\end{figure*}

\begin{figure*}
   \centering
   \includegraphics[width=1\linewidth]{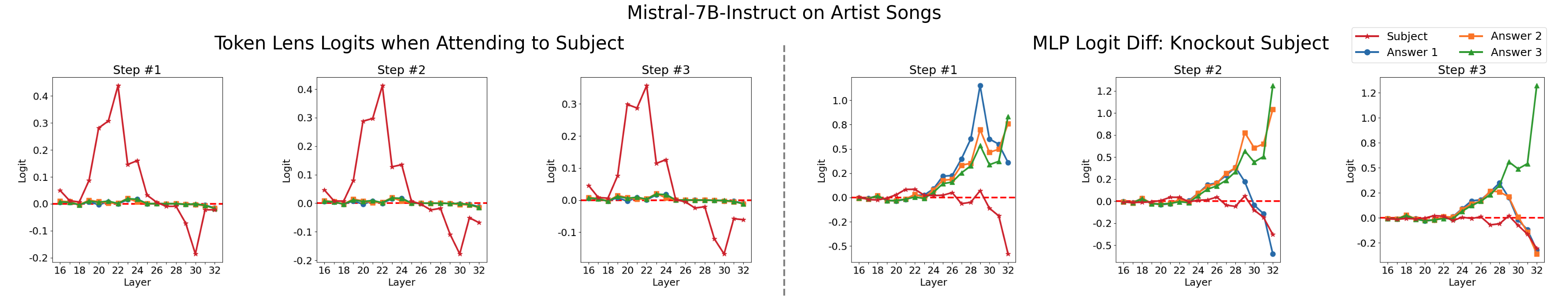}
   \caption{Token Lens logit values (left) and MLP logit differences (right) of subject and answer tokens of Mistral-7B-Instruct on Artist-Songs dataset when attending to or knocking out the subject tokens (averaged across three prompt templates).}
   \label{fig:artist_songs-mistral-subject}
\end{figure*}

\begin{figure*}
   \centering
   \includegraphics[width=1\linewidth]{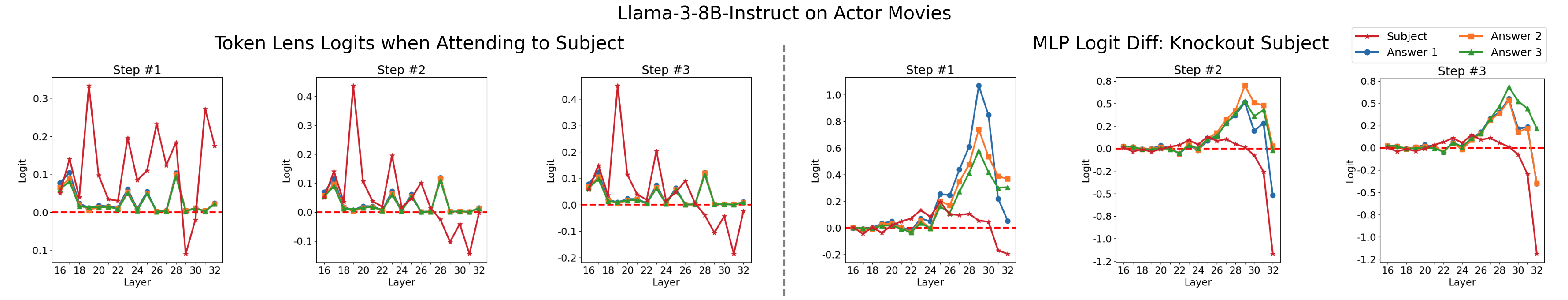}
   \caption{Token Lens logit values (left) and MLP logit differences (right) of subject and answer tokens of Llama-3-8B-Instruct on Actor-Movies dataset when attending to or knocking out the subject tokens (averaged across three prompt templates).}
   \label{fig:actor_movies-llama-subject}
\end{figure*}

\begin{figure*}
   \centering
   \includegraphics[width=1\linewidth]{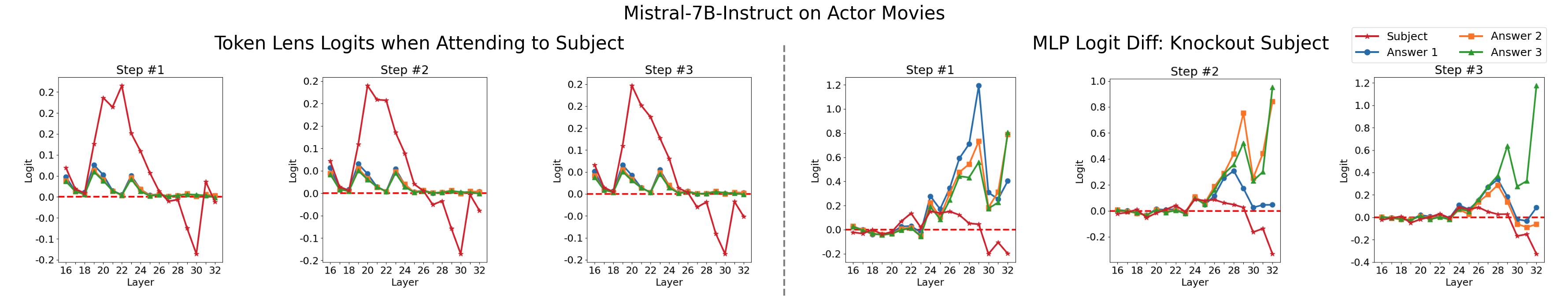}
   \caption{Token Lens logit values (left) and MLP logit differences (right) of subject and answer tokens of Mistral-7B-Instruct on Actor-Movies dataset when attending to or knocking out the subject tokens (averaged across three prompt templates).}
   \label{fig:actor_movies-mistral-subject}
\end{figure*}

\Cref{fig:token_lens_logit_attention_to_subject}, \Cref{fig:attention_knockout_subject_logits}, \Cref{fig:token_lens_logit_attention_to_prev_ans}, \Cref{fig:attention_knockout_prev_ans_logits}, \Cref{fig:token_lens_last_token_logits}, \Cref{fig:attention_knockout_last_token_logits} are the complete results for Token Lens and Attention Knockout analyses on the subject token, previous answer tokens, and the last token. The results are macro-averaged across three answer steps and aggregated over all models and datasets. \Cref{fig:country_cities-llama-subject}, \Cref{fig:country_cities-mistral-subject}, \Cref{fig:artist_songs-llama-subject}, \Cref{fig:artist_songs-mistral-subject}, \Cref{fig:actor_movies-llama-subject}, \Cref{fig:actor_movies-mistral-subject} are the Token Lens and attention Knockout results on the subject token from different models and datasets. The pattern of attention using the subject token to promote answers is more prominent in the Country-Cities dataset (\Cref{fig:country_cities-llama-subject}, \Cref{fig:country_cities-mistral-subject}) compared to the other two datasets (\Cref{fig:artist_songs-llama-subject}, \Cref{fig:artist_songs-mistral-subject}, \Cref{fig:actor_movies-llama-subject}, \Cref{fig:actor_movies-mistral-subject}). Refer to \url{https://drive.google.com/drive/folders/1HtMtgm63ZZDfAnjeFDLJqMwSvLSyWAlj?dmr=1&ec=wgc-drive-globalnav-goto} for dataset- and model-specific figures on all different tokens without early layers omitted. 

\section{Analysis on More Answer Steps}
\label{appendix_more_answer_steps}

\subsection{Five Answer Steps}
We asked the models to generate five object entities with prompt template $1$. However, only the Actor-Movies dataset yielded over $100$ correct cases from both models. The other datasets did not meet this threshold as model performance declined with more answer steps. See \Cref{tab:model_performance_on_five_answer_steps} for the accuracy of each model on every dataset.

\renewcommand{\arraystretch}{1.3}
\begin{table}[ht]
\centering
\scriptsize
\begin{tabular}{c|c|c}
\toprule
\textbf{Dataset} & \textbf{Llama-3-8B-Instruct}& \textbf{Mistral-7B-Instruct-v0.2}\\
\hline
Country-Cities& 92/167 (55.1\%)& 87/167 (52.1\%)\\
\hline
Artist-Songs& 82/2076 (4.0\%)& 74/2076 (3.6\%)\\
\hline
Actor-Movies& 582/7914 (7.4\%)& 422/7914 (5.3\%)\\
\bottomrule
\end{tabular}
\caption{Number of correct cases and accuracy of two models on each dataset when the number of object entity $n=5$.}
\label{tab:model_performance_on_five_answer_steps}
\end{table}

We conducted token-level analyses with methods described in \Cref{6_1_token_analysis_methodology}. We found that all major results and patterns at answer steps $4$ and $5$ match with those from answer steps $1$, $2$, and $3$:

\begin{figure*}
    \centering
    \includegraphics[width=1\linewidth]{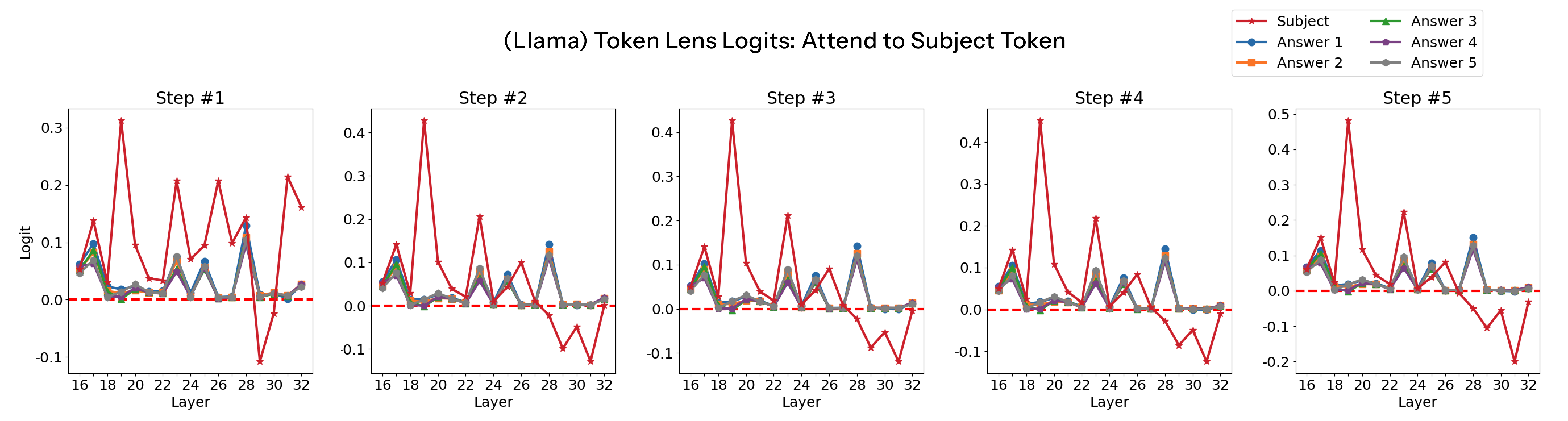}
    \caption{Llama's Token Lens logit values of subject and answer tokens across layers and answer steps when attending to the subject (macro-averaged across all datasets with template 1). Attention promotes and extracts subject information in the middle layers while suppressing it in later layers, which is consistent across all five answer steps.}
    \label{fig:five_answer_step-llama-subject-token_lens}
\end{figure*}

\begin{figure*}
    \centering
    \includegraphics[width=1\linewidth]{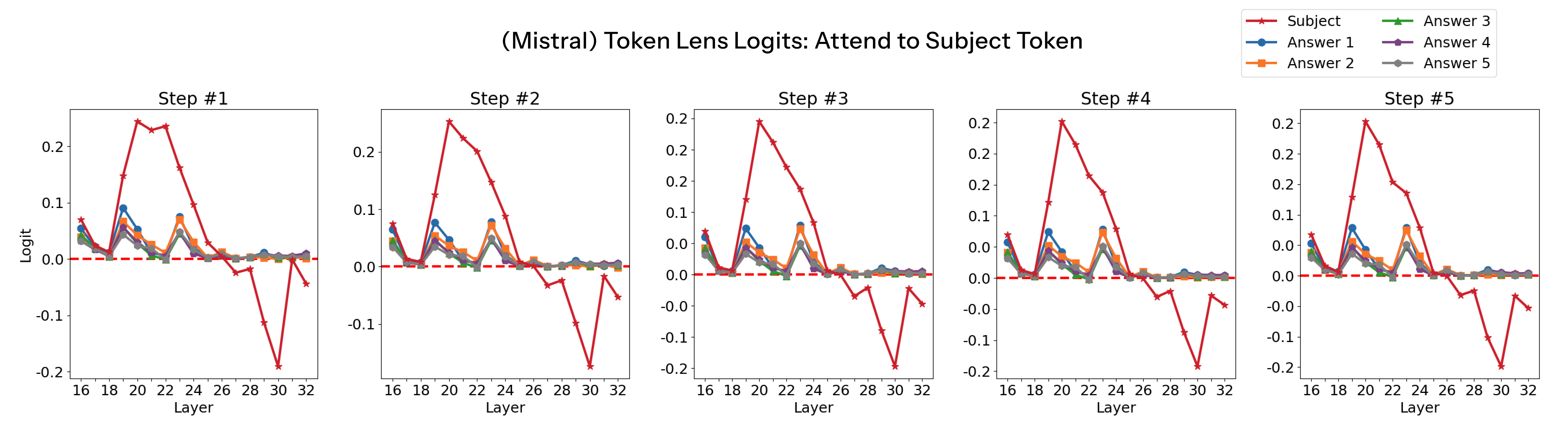}
    \caption{Mistral's Token Lens logit values of subject and answer tokens across layers and answer steps when attending to the subject (macro-averaged across all datasets with template 1). The patterns at steps 4 and 5 align with those observed in the first three steps.}
    \label{fig:five_answer_step-mistral-subject-token_lens}
\end{figure*}

\begin{figure*}
    \centering
    \includegraphics[width=1\linewidth]{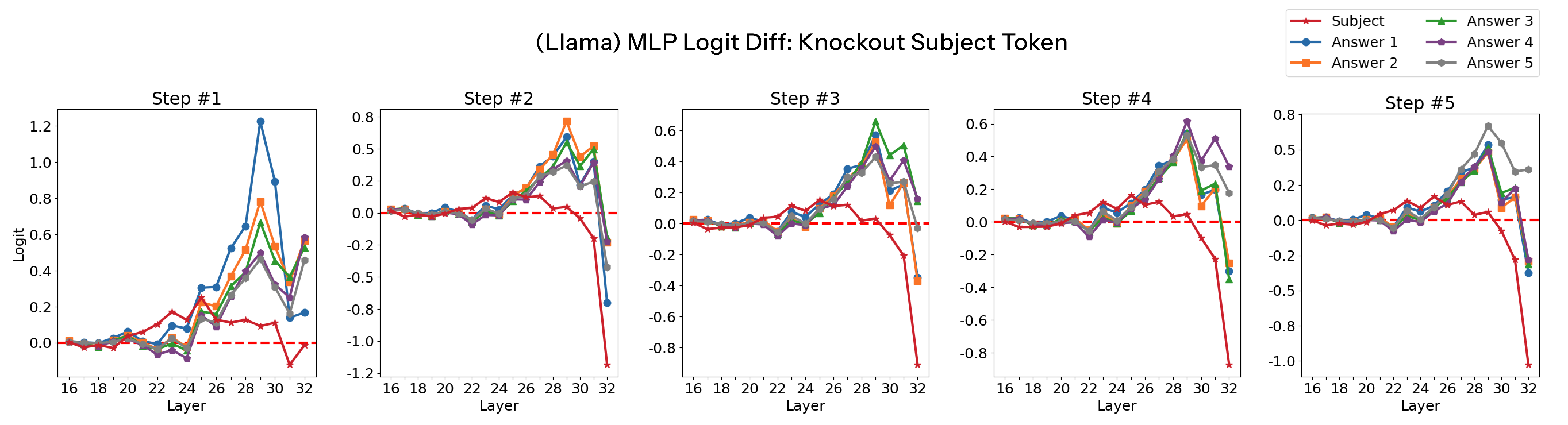}
    \caption{Llama's Logit differences of the subject and answer tokens between MLP outputs with and without knocking out attention from the last to the subject tokens (macro-averaged across all datasets with template 1). Positive logit differences for the answers and negative differences for the subject in later layers show that MLPs use the subject information to promote answers and suppress the subject. This pattern is consistent across all five answer steps.}
    \label{fig:five_answer_step-llama-subject-attn_knockout}
\end{figure*}

\begin{figure*}
    \centering
    \includegraphics[width=1\linewidth]{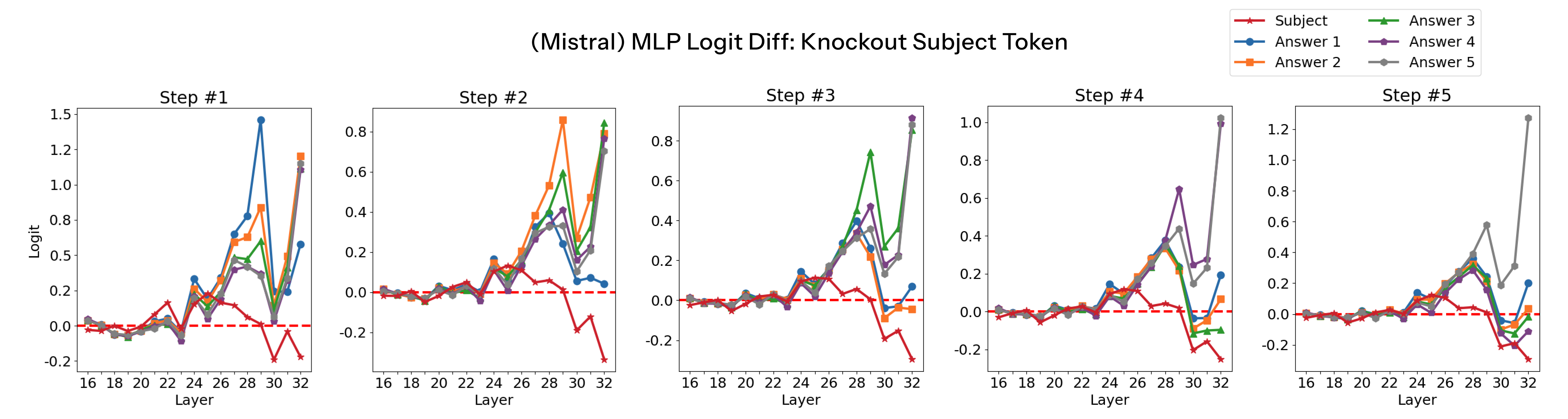}
    \caption{Mistral's logit differences of the subject and answer tokens between MLP outputs with and without knocking out attention from the last to the subject tokens (macro-averaged across all datasets with template 1). The patterns at steps $4$ and $5$ align with those observed from the first three steps.}
    \label{fig:five_answer_step-mistral-subject-attn_knockout}
\end{figure*}


\begin{figure*}
    \centering
    \includegraphics[width=1\linewidth]{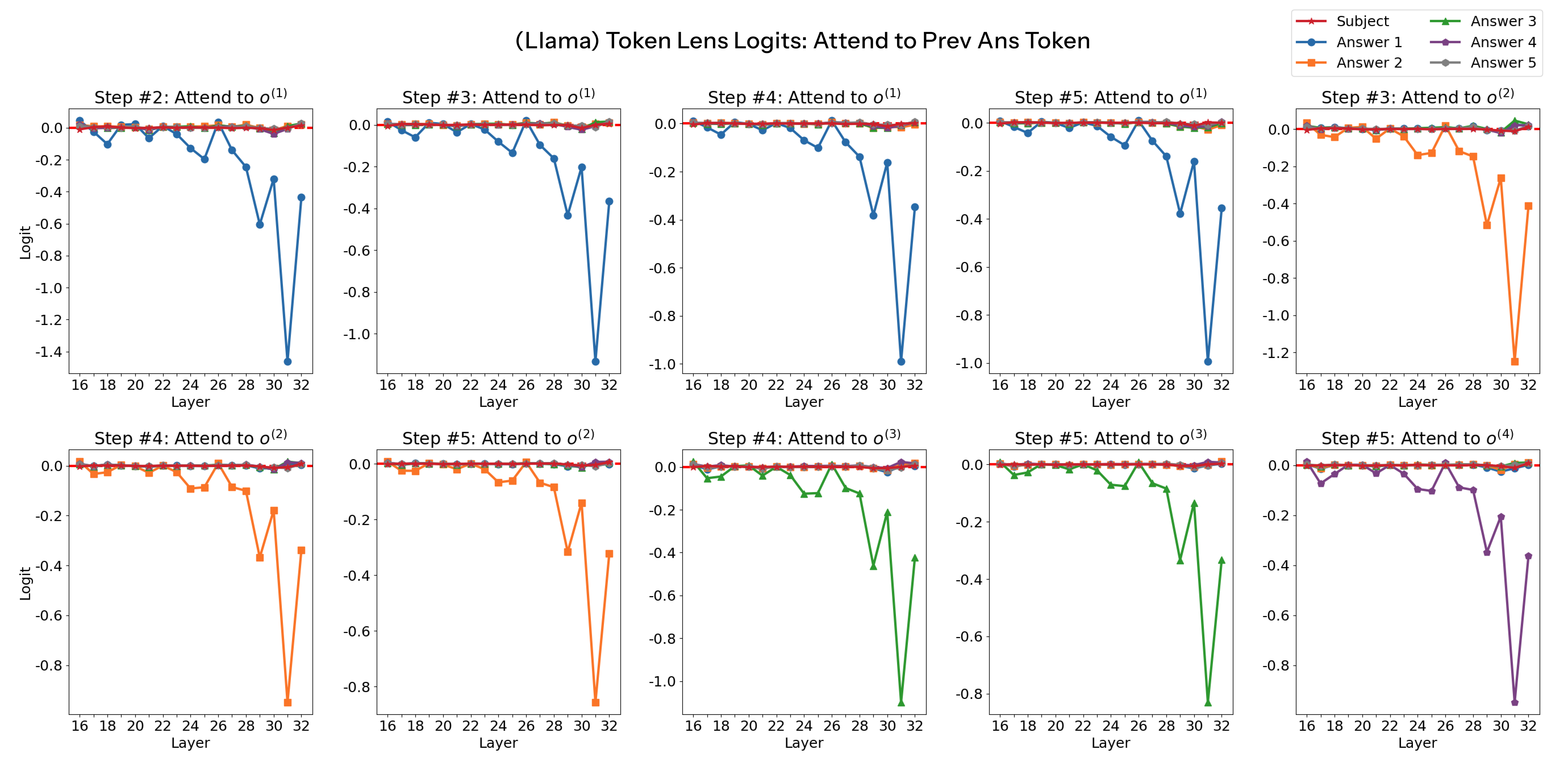}
    \caption{Llama's Token Lens logit values subject and answer tokens across layers and answer steps (macro-averaged across all datasets, models, and templates) when attending to previous answers. The logit of the attended answer is negative at later layers, showing that the attention is suppressing previously generated answers. The patterns at answer step $4$ and $5$ match with the ones discussed in the main section.}
    \label{fig:five_answer_steps-llama-prev_ans-token_lens}
\end{figure*}

\begin{figure*}
    \centering
    \includegraphics[width=1\linewidth]{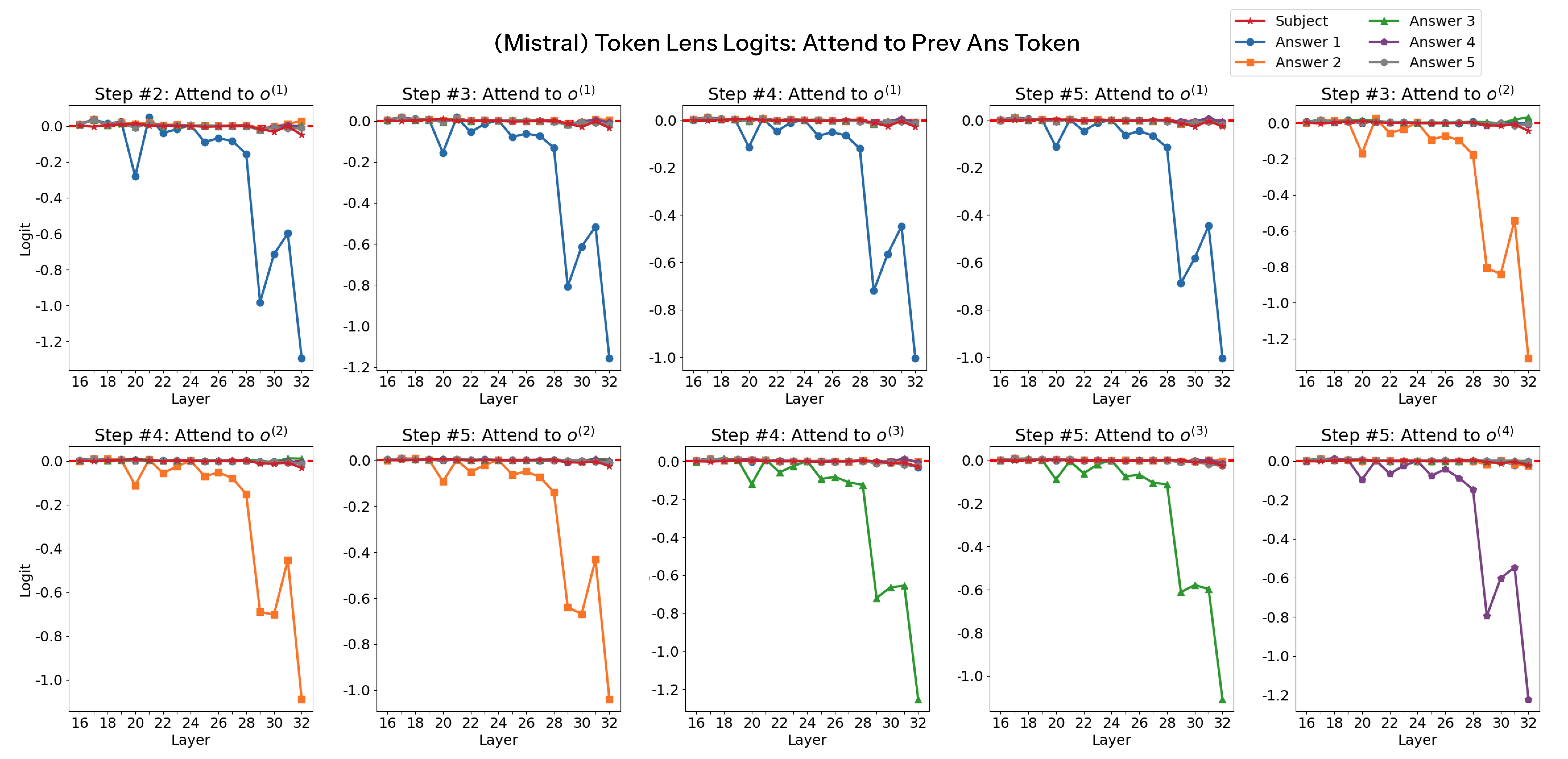}
    \caption{Mistral's Token Lens logit values subject and answer tokens across layers and answer steps (macro-averaged across all datasets, models, and templates) when attending to previous answers. The logit of the attended answer is negative at later layers, showing that the attention is suppressing previously generated answers. The patterns at steps 4 and 5 align with those observed in Llama and in the first three steps.}
    \label{fig:five_answer_steps-mistral-prev_ans-token_lens}
\end{figure*}

\begin{figure*}
    \centering
    \includegraphics[width=1\linewidth]{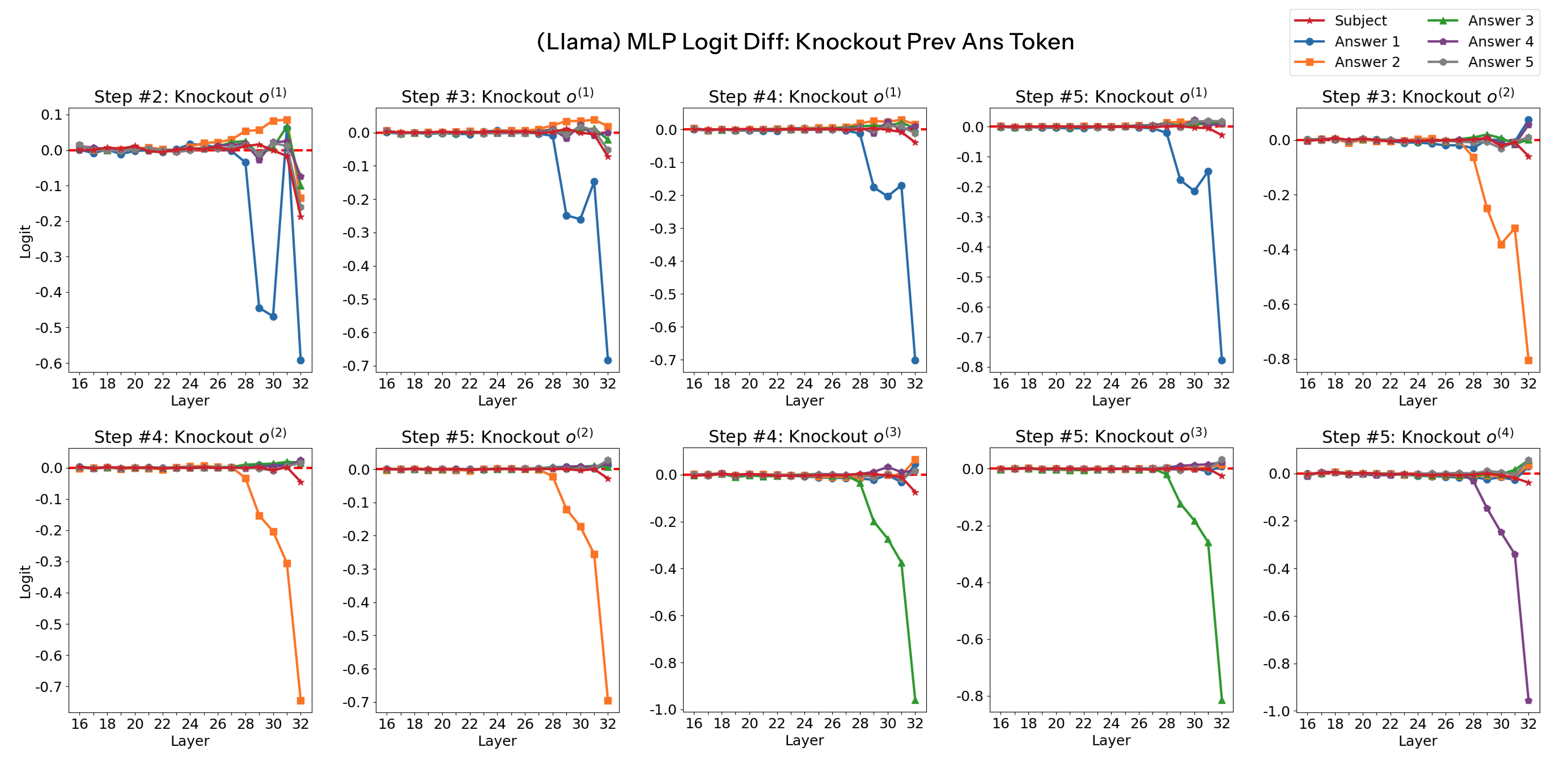}
    \caption{Llama's Logit differences for subject and answer tokens between MLP outputs with and without knocking attention from the last to previous answer tokens (macro-averaged across all datasets, models, and templates). All previously generated answer tokens have negative logits, and all new answers have positive logits. This result suggests that MLPs use previous answers for both repetition suppression and new answer promotion, aligning with the patterns discussed in the main section.}
    \label{fig:five_answer_steps-llama-prev_ans-attn_knockout}
\end{figure*}

\begin{figure*}
    \centering
    \includegraphics[width=1\linewidth]{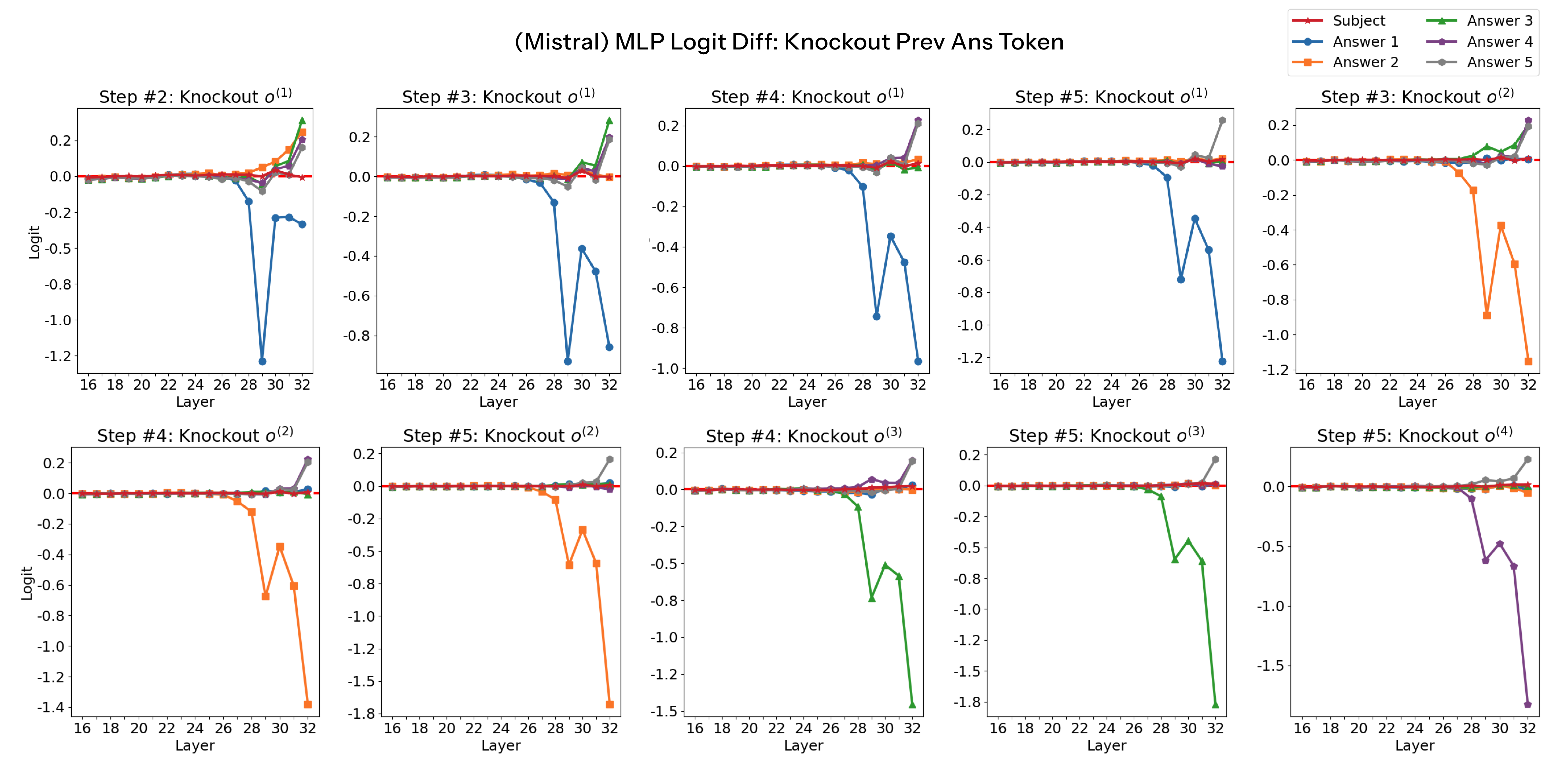}
    \caption{Mistral's logit differences for subject and answer tokens between MLP outputs with and without knocking attention from the last to previous answer tokens (macro-averaged across all datasets, models, and templates). The patterns at steps $4$ and $5$ align with those observed in the first three steps.}
    \label{fig:five_answer_steps-mistral-prev_ans-attn_knockout}
\end{figure*}

\begin{figure*}
    \centering
    \includegraphics[width=1\linewidth]{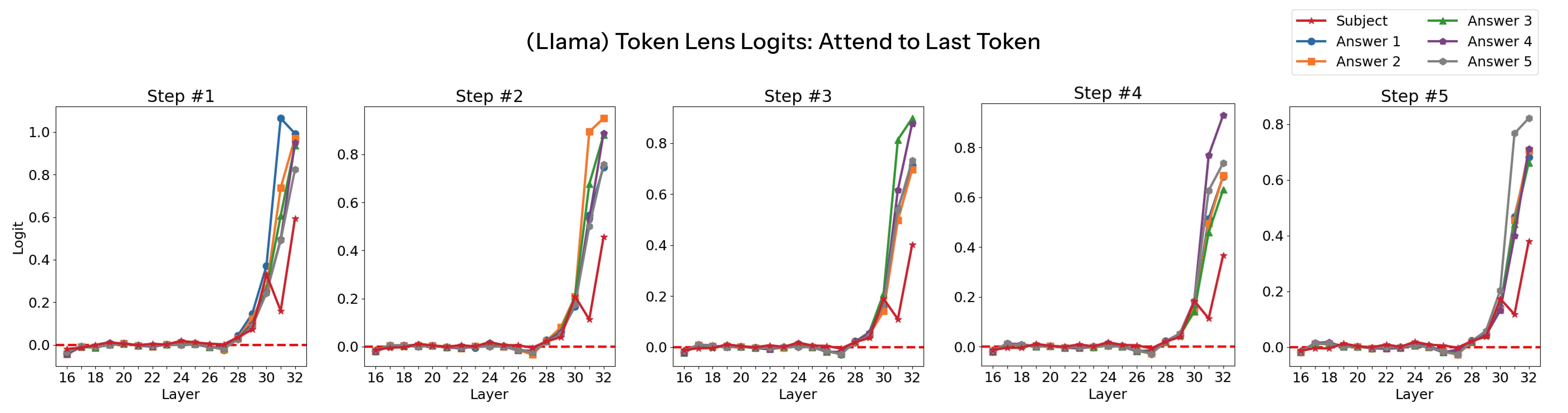}
    \caption{Llama's Token Lens logit values of subject and answer tokens across layers and answer steps when attending to the last token (macro-averaged across all datasets, models, and templates). Attention promotes all three answers and the subject at the final layers, with the answer for the current step having the highest logit, which align with the findings discussed in the main section.}
    \label{fig:five_answer_steps-llama-last_token-token_lens}
\end{figure*}

\begin{figure*}
    \centering
    \includegraphics[width=1\linewidth]{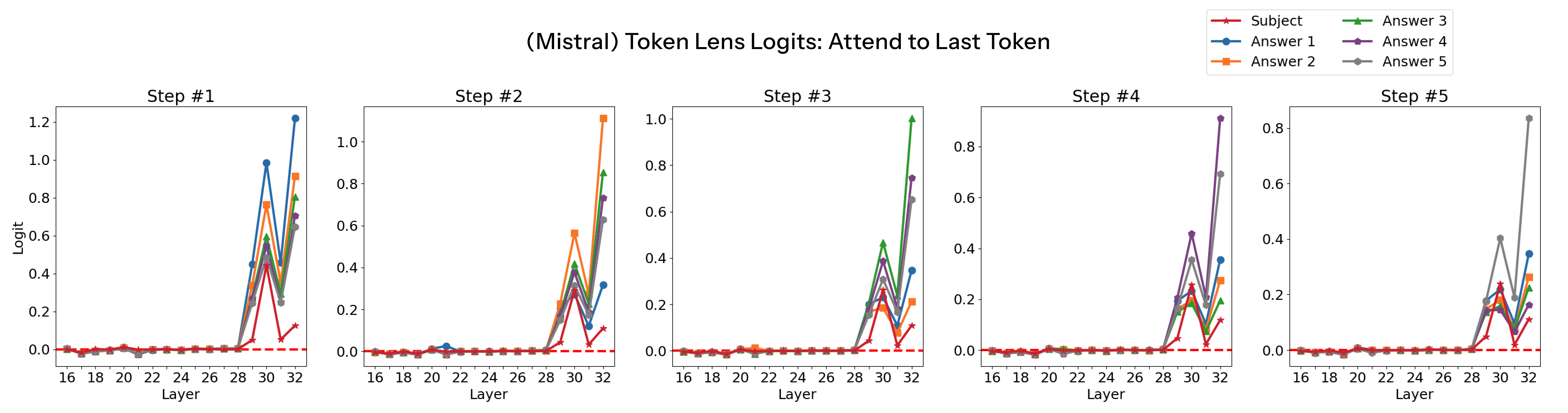}
    \caption{Mistral's Token Lens logit values of subject and answer tokens across layers and answer steps when attending to the last token (macro-averaged across all datasets, models, and templates). The patterns at steps $4$ and $5$ align with those observed in the first three steps.}
    \label{fig:five_answer_steps-mistral-last_token-token_lens}
\end{figure*}

\begin{figure*}
    \centering
    \includegraphics[width=1\linewidth]{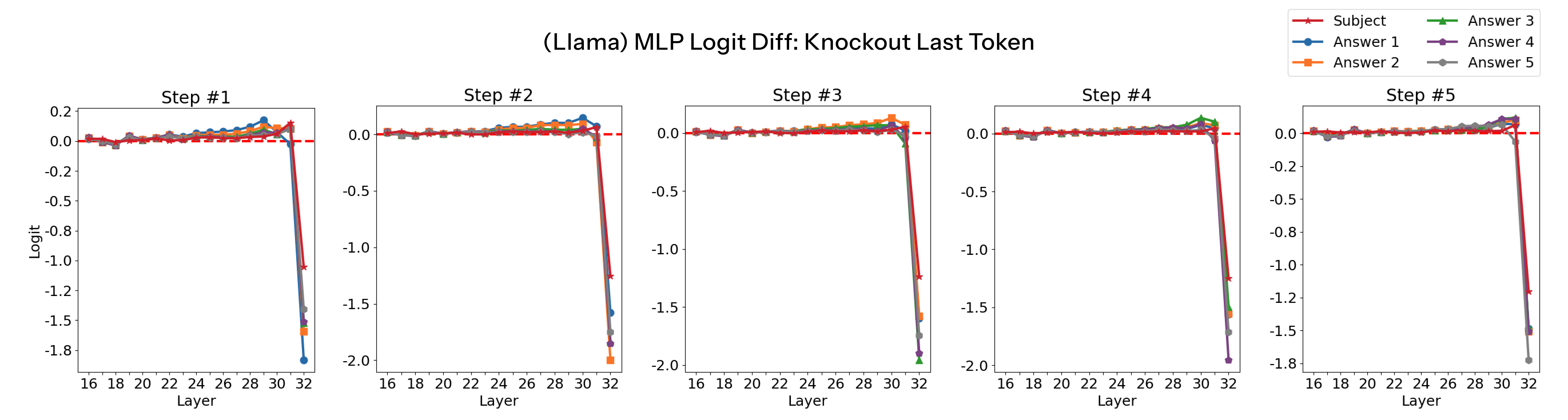}
    \caption{Llama's logit differences for subject and answer tokens between MLP outputs with and without knocking attention from the last token to itself (macro-averaged across all datasets, models, and templates). The pattern is aligned with those discussed in the main section.}
    \label{fig:five_answer_steps-llama-last_token-attn_knockout}
\end{figure*}

\begin{figure*}
    \centering
    \includegraphics[width=1\linewidth]{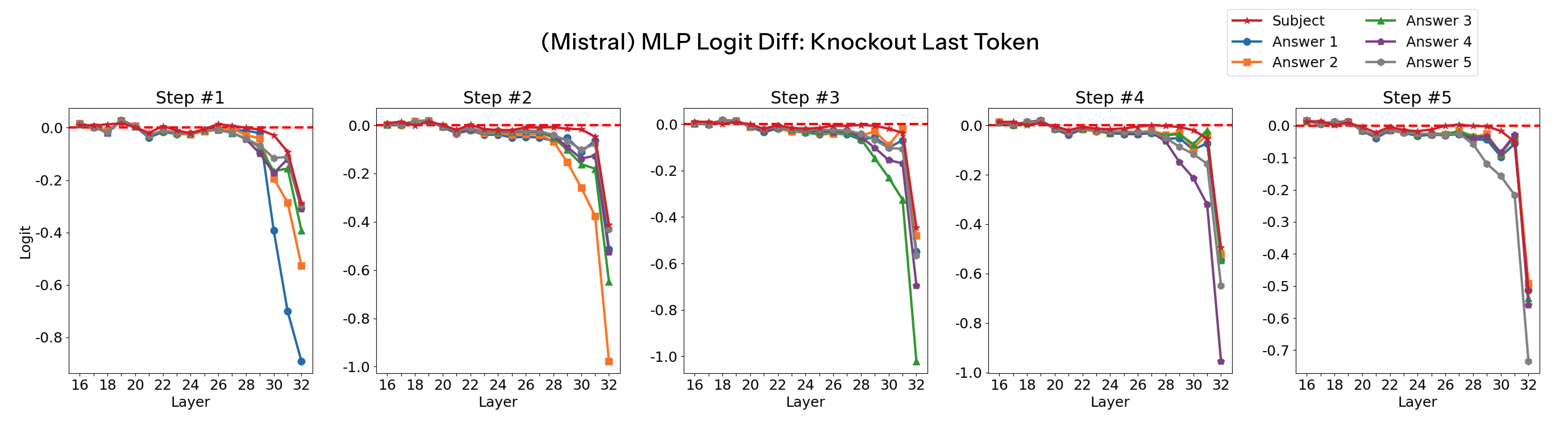}
    \caption{Mistral's logit differences for subject and answer tokens between MLP outputs with and without knocking attention from the last token to itself (macro-averaged across all datasets, models, and templates). The patterns at steps $4$ and $5$ align with those observed in the first three steps.}
    \label{fig:five_answer_steps-mistral-last_token-attn_knockout}
\end{figure*}

\begin{itemize}
    \item Attention attends to subject tokens, promoting them in middle layers and suppressing them in deeper layers (\Cref{fig:five_answer_step-llama-subject-token_lens}, \Cref{fig:five_answer_step-mistral-subject-token_lens}). MLPs use the subject to promote answers at the middle layers (\Cref{fig:five_answer_step-llama-subject-attn_knockout}, \Cref{fig:five_answer_step-mistral-subject-attn_knockout}).
    
    \item Attention also attends to previous answers to suppress them (\Cref{fig:five_answer_steps-llama-prev_ans-token_lens}, \Cref{fig:five_answer_steps-mistral-prev_ans-token_lens}), with MLPs reinforcing this suppression and promoting new answers in later layers (\Cref{fig:five_answer_steps-llama-prev_ans-attn_knockout}, \Cref{fig:five_answer_steps-mistral-prev_ans-attn_knockout}).
    
    \item Attention at the last token promotes answers in the final layers (\Cref{fig:five_answer_steps-llama-last_token-token_lens}, \Cref{fig:five_answer_steps-mistral-last_token-token_lens}), and MLPs compensate answer promotion when direct attention to the last token is intervened (\Cref{fig:five_answer_steps-llama-last_token-attn_knockout}, \Cref{fig:five_answer_steps-mistral-last_token-attn_knockout}).
\end{itemize}

\subsection{Ten Answer Steps}
We also tried with $10$ answer steps, but we could not collect $100$ correct cases from any model and dataset. The model performance became much worse. For example, among all $154$ Country-Cities data entries with at least $10$ answers, Llama-3-8B-Instruct only got $56$ correct, and Mistral-7B-Instruct-v0.2 only got $45$ correct.

\end{document}